\documentclass[a4paper,fleqn]{cas-dc}

\usepackage[authoryear]{natbib}
\usepackage{multirow}
\usepackage{booktabs} 
\usepackage{makecell}
\usepackage{soul}
\usepackage{bbm}
\usepackage{array}
\usepackage{threeparttable}
\usepackage[ruled,vlined]{algorithm2e}
\usepackage{subcaption}
\usepackage{caption}
\usepackage[table]{xcolor}
\usepackage{adjustbox}

\def\tsc#1{\csdef{#1}{\textsc{\lowercase{#1}}\xspace}}
\tsc{WGM}
\tsc{QE}

\begin{document}
\let\printorcid\relax  
\let\WriteBookmarks\relax
\def\floatpagepagefraction{1}
\def\textpagefraction{.001}

\shorttitle{SurgSLOT: Segment Anything in Surgical Videos}   

\shortauthors{H. Liu et al.}  

\title [mode = title]{SurgSLOT: Segment Anything in Surgical Videos via Semantic Long-term Tracking}  

\author[1]{Haofeng Liu}
\credit{Conceptualization, Methodology, Software, Investigation, Data curation, Writing – original draft \& editing}

\author[1]{Ziyue Wang}
\credit{Conceptualization, Validation, Writing – review \& editing}

\author[1]{Sudhanshu Mishra}
\credit{Data curation, Validation, Writing – review \& editing}

\author[4]{Mingqi Gao}
\credit{Formal analysis, Writing – review}

\author[1]{Guanyi Qin}
\credit{Writing – review \& editing}

\author[1]{Chang~Han Low}
\credit{Data curation, Writing – review}

\author[1]{Alex~Y.~W. Kong}
\credit{Validation, Software}

\author[1]{Zhu Zhuo}
\credit{Software}

\author[3]{Huazhu Fu}
\credit{Writing – review \& editing}

\author[2]{Joseph S. Ng}
\credit{Validation, Data curation}

\author[1]{Yueming Jin}
\corref{cor1}
\cortext[cor1]{Corresponding author.}
\ead{ymjin@nus.edu.sg}
\credit{Supervision, Project administration, Conceptualization, Funding acquisition, Writing – review \& editing}

\affiliation[1]{
    organization={National University of Singapore},
    city={Singapore},
    country={Singapore}
}


\affiliation[2]{
    organization={National University Hospital},
    city={Singapore},
    country={Singapore}
}

\affiliation[3]{
    organization={Institute of Advanced Intelligence and Computing, Agency for Science, Technology and Research},
    city={Singapore},
    country={Singapore}
}

\affiliation[4]{
    organization={The University of Sheffield},
    city={Sheffield},
    country={UK}
}

\nonumnote{Code and dataset will be released at https://jinlab-imvr.github.io/SurgSLOT/.}
\nonumnote{This work was supported by the Ministry of Education, Singapore, under the Tier 2 grant (T2EP20224-0028) and the Tier 1 grant (23-0651-P0001).}

\begin{abstract}
Surgical scene understanding demands temporally consistent tracking of instruments and tissues.
For clinical use, such tracking should generalize to new centers and procedure types, yet retraining for each of them is costly and not scalable.
Interactive video object segmentation offers a way toward this generalization: the target is specified at inference by a first-frame visual prompt, so a model can generalize to unseen categories and new scenarios without retraining.
However, training such a generalizable model demands spatio-temporal masklet annotations at a scale and procedural diversity that existing surgical benchmarks lack.
We fill this gap with iSurg, the largest surgical segmentation benchmark to our knowledge, spanning six procedure types with over $170$k frames, $410$k object masks, and $2.4$k masklets, including an \textit{in-house clinical dataset} of four $30$-minute videos.
Yet tracking over such long videos remains challenging: the target needs to be re-identified among visually similar objects after long absences, while long-term memory has to judge whether each stored frame genuinely depicts the target, both demanding a stable object-level semantic identity.
To this end, we propose SurgSLOT, a surgical segmentation generalist that segments any prompted target through two coupled modules built on this semantic identity:
Temporal Semantic Learning learns it for re-identification, and Semantic-driven Long-term Memory reuses it to select reliable memory frames, suppressing identity drift over long procedures.
On the SAM2 and SAM3 backbones, SurgSLOT reaches $81.0$ and $82.8$ Macro Average $\mathcal{J}$\&$\mathcal{F}$ under cross-dataset evaluation, surpassing their fine-tuned counterparts by $5.1$ and $5.3$ points and transferring zero-shot to an unseen procedure type and unseen object categories, with the SAM2 version running in real time at $68$ FPS.
\end{abstract}


\maketitle

\section{Introduction}
\label{sec:introduction}
\begin{figure}
    \centering
    \includegraphics[width=\linewidth]{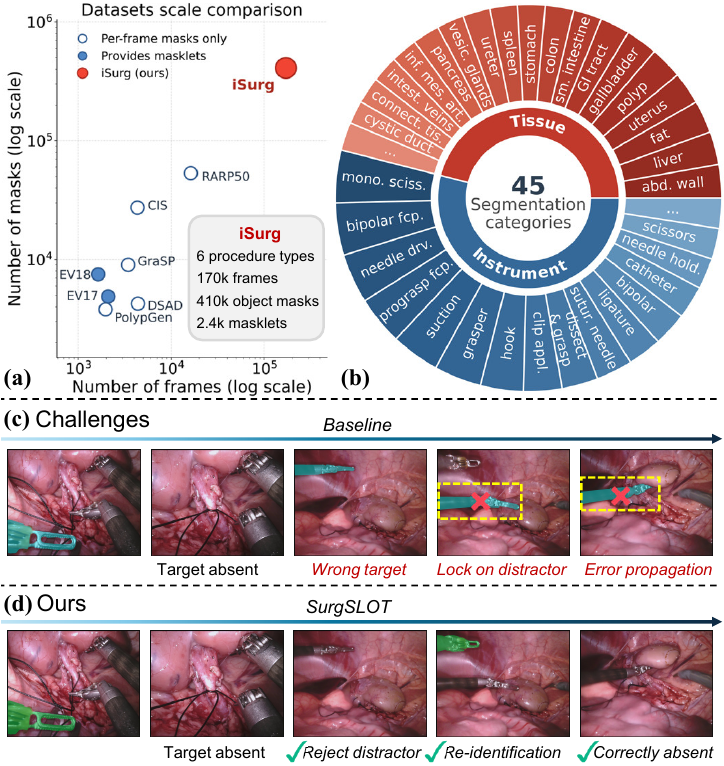}
    \caption{Overview of the iSurg benchmark and the challenges motivating SurgSLOT.
    (a) iSurg is the largest surgical segmentation benchmark spanning multiple procedure types with masklet annotations.
    (b) Per-category mask distribution over $20$ instrument and $25$ tissue categories.
    (c, d) The same surgical video tracked by a baseline (c) and by SurgSLOT (d) over a long procedure: the baseline drifts as the target disappears and a similar instrument appears, whereas SurgSLOT re-identifies the target and maintains consistent tracking.}
    \label{fig:pie}
\end{figure}

Surgical video segmentation localizes and tracks instruments and anatomical structures across video sequences, serving as a foundational perception capability for computer-assisted surgery \citep{du2019pba, jin2022exploring}.
During surgery, it enables context-aware interpretation of the scene for cognitive assistance, surgeon perception, operating workflow optimization, and intraoperative decision support \citep{ahmed2024deep}.
Beyond the operating room, it supports post-operative tasks such as surgeon skill assessment, report documentation, and surgical education \citep{maier2022surgical}.
More fundamentally, instance-level segmentation and tracking of instruments and tissues underpin downstream applications such as tool pose estimation, trajectory prediction, and even surgical robot autonomy \citep{du2019pba, saeidi2022autonomous, long2025surgical}.
In practice, however, these applications demand more than accuracy on a single dataset.
Surgical videos differ across clinical centers due to variations in recording systems and imaging protocols, and across procedure types due to differences in anatomy and instruments.
Even within the same institution, considerable appearance variations may arise from differences in patient anatomy, intra-operative conditions (e.g., bleeding or smoke), and surgeon operative skills.
Therefore, an ideal clinically applicable model should generalize beyond a single dataset to unseen clinical centers and procedure types, toward the goal of \textit{segmenting anything in surgical videos} across diverse and unseen scenarios.

However, most surgical segmentation studies are still developed and evaluated within a single dataset, with limited generalization to new clinical centers and procedure types \citep{ni2020feature, gonzalez2020in, wang2025lacoste}.
One obvious remedy is to retrain a dedicated model for each new context, but collecting and annotating data at every site is time-consuming and expensive, demands clinical expertise, and does not scale to real-world deployment.
Only a few works pursue generalization explicitly. Test-time online adaptation transfers an instrument segmenter to a new video using a single annotated frame \citep{zhao2021anchor}, while single-domain generalization learns a domain-adaptive segmenter that transfers to unseen clinical centers and devices \citep{guo2024infproto}. 
Yet the first fine-tunes the model on every new video at test time, and the second stays bound to the categories fixed at training, unable to segment a category absent from training, no matter how large the training set.
A different route avoids both limitations: it specifies the target at inference with a first-frame visual prompt (point, box, or mask) and propagates it throughout the video, requiring no test-time training and no fixed category set.
Interactive video object segmentation (iVOS) realizes this through promptable segmentation foundation models such as SAM2 \citep{sam2} and SAM3 \citep{sam3}, which segment and track any target from a visual prompt and have shown generalization in the natural domain.
Surgical adaptations have begun to adopt it \citep{surgicalsam2, masam2}, yet remain confined to single-dataset development, leaving their cross-dataset generalization underexplored.
Given the clinical importance of surgical generalization, we pursue it within this promptable paradigm built on segmentation foundation models.

Building such a generalizable model requires large-scale and diverse data that existing surgical benchmarks do not provide.
Because tracking a prompted target across all subsequent frames relies on annotations that follow the same object through time, it requires masklets \citep{sam2}: temporally consistent sequences of masks, each tracking one target throughout a video, instead of isolated per-frame labels.
Such masklets are scarce in surgery. Most existing datasets provide only per-frame semantic masks, and the few reorganized into masklets remain limited in scale and cover only a single procedure type \citep{resurgsam2}, leaving cross-dataset generalization neither learnable nor reliably measurable.
We take the first step toward filling this gap with iSurg (Interactive Segmentation in Surgical Videos), illustrated in Fig.~\ref{fig:pie}, the largest surgical segmentation benchmark to our knowledge. It provides the scale and procedural diversity that cross-dataset generalization requires, spanning six procedure types with $735$ videos totaling over $43$ hours, $170$k frames, $410$k object masks, and $2.4$k masklets.
Built by reconstructing per-frame annotations into masklets and scaling masklets through a self-evolving data engine, iSurg is split for \textbf{cross-dataset generalization} \citep{huang2025eval}: every test set is drawn from a data source unseen during training. 
Two stricter zero-shot settings further evaluate transfer to an unseen procedure type and to unseen object categories.
Within the test split, we further curate an in-house dataset (NUH-Hyst) of four clinical hysterectomy videos, each up to $30$ minutes, held out for long-term evaluation.

Beyond the data foundation, deploying such a model on surgical videos raises a central methodological challenge: a target needs to be tracked over sequences that span tens of minutes, far longer than the short clips of general-domain video. 
Tracking over such long sequences brings two difficulties (Fig.~\ref{fig:pie}(c,d)).
First, the target repeatedly leaves and re-enters the field of view, and on reappearance, it needs to be re-identified and distinguished from visually similar objects in the scene.
Second, long-term tracking relies on memory: it stores earlier frames so the target can be recovered when it reappears. The difficulty is judging whether each stored frame genuinely depicts the target, a judgment that mask quality alone cannot reliably provide.
Existing methods keep frames by mask quality~\citep{masam2, dam4sam}, so a visually similar distractor segmented as a clean, confident mask can be admitted to memory and propagated as a reference onto the wrong object, compounding into progressive drift.
Re-identification and reliable memory thus share one underlying need: a stable object-level identity for the tracked target.
Such an identity emerges by aggregating the target over time, capturing the semantics that persist across frames instead of the momentary appearance of any single one.
These persistent semantics are less tied to any single dataset, so the same identity that stabilizes long-term tracking can also support transfer across clinical centers, procedure types, and object categories unseen during training.

Building on iSurg, we propose SurgSLOT, a \textbf{Surg}ical segmentation generalist that segments any prompted target via \textbf{S}emantic \textbf{LO}ng-term \textbf{T}racking.
It couples target re-identification and long-term memory selection through a single object identity that stays reliable throughout extended surgical procedures.
To build this identity, we propose Temporal Semantic Learning (TSL), which learns a per-frame semantic representation and aggregates a target's reliable appearances into a slowly evolving semantic anchor.
Pulling reliable appearances of the same target toward this shared anchor, and pushing away frames where it is absent or wrongly localized, encourages the semantic representation to capture what stays invariant across viewpoint, occlusion, and motion, so a target reappearing after a long absence still aligns with its anchor and is re-identified.
Vision-language contrastive learning complements this by separating instruments of different categories, reducing confusion among visually similar ones. 
Beyond re-identification, the same identity guides memory selection: we design a Semantic-driven Long-term Memory (SLM) that keeps reliable frames from much earlier in the video.
Attending to such distant frames requires two components. Long-range sampling trains the memory attention to associate references across long temporal gaps, which standard short-clip training does not cover. Anchor-aligned selection reuses the TSL anchor to admit only frames whose semantics stay consistent with it and reject visually similar distractors whose semantics diverge, storing frames by identity instead of by mask quality.
In this way, SurgSLOT tracks targets consistently over long procedures, recovering them after long absences and suppressing memory contamination and drift.

The main contributions of this paper are as follows:
\begin{itemize}
\item We establish iSurg, the largest surgical segmentation benchmark to our knowledge, providing masklet annotations across six procedure types. 
It is built through dataset reconstruction and a self-evolving data engine, with an in-house clinical dataset for long-term evaluation, and split for cross-dataset generalization, including two zero-shot settings on an unseen procedure type and unseen object categories.
\item We propose SurgSLOT, a surgical segmentation generalist. Its Temporal Semantic Learning builds an object-level semantic anchor as a stable target identity, enabling re-identification after long absences and transfer to unseen procedures and categories.
\item We further design Semantic-driven Long-term Memory, which selects the long-term memory frames most consistent with this anchor, suppressing the identity drift that accumulates over long surgical videos.
\item Across diverse surgical datasets, SurgSLOT achieves state-of-the-art cross-dataset performance, generalizing to an unseen procedure type and to unseen object categories, with consistent gains on both SAM2 and SAM3 backbones.
\end{itemize}

\section{Related Work}
\subsection{Interactive Video Object Segmentation}
Video Object Segmentation (VOS) aims to segment and track objects across video sequences \citep{cheng2022xmem, gao2023deep}. Beyond semi-supervised VOS requiring complete first-frame masks \citep{cutie, qin2025structure}, iVOS enables users to specify targets with lightweight prompts such as points or boxes \citep{wong2024scribbleprompt}. 
SAM2 and its successor SAM3 are state-of-the-art segmentation foundation models that couple promptable segmentation with memory-based temporal modeling, enabling iVOS that tracks a visually prompted target across frames through propagation.
However, directly applying these natural-video models to surgery falls short under cross-dataset deployment, where prolonged procedures and visually similar instruments jointly undermine robustness.

\subsection{Surgical Video Segmentation}
Surgical video segmentation has progressed from closed-set per-frame prediction toward foundation-model-based interactive segmentation. 
Early methods predict per-pixel semantic masks over predefined categories \citep{ni2020feature}, later extended to per-frame instances \citep{gonzalez2020in} and to Transformer query formulations with temporal or stereo context \citep{zhao2022trasetr, ayobi2023matis, wang2025lacoste}. Even foundation-model-based variants such as SurgicalSAM \citep{Yue2024surgsam} stay within this paradigm, committing to a fixed label space and yielding per-frame predictions rather than temporally consistent masklets.
Interactive segmentation instead decouples target specification from category supervision through prompts \citep{kirillov2023sam, ma2024segment, sam2}. 
Video-level adaptations such as SurgicalSAM2 \citep{surgicalsam2}, MedSAM2 \citep{medsam2}, MA-SAM2 \citep{masam2}, and ReSurgSAM2 \citep{resurgsam2} extend SAM2 to propagate masklets across surgical videos.

A separate line of work targets cross-domain transfer directly: test-time adaptation transfers an instrument segmenter to a new video from one annotated frame \citep{zhao2021anchor}, and single-domain generalization learns a domain-adaptive segmenter from one source domain to reach unseen centers and devices \citep{guo2024infproto}.
The former fine-tunes the model on each new video at test time, and the latter stays restricted to a fixed category set. In contrast, iVOS transfers to held-out datasets from a first-frame prompt with no test-time training, specifying the target at inference without a fixed category set.
Yet whether iVOS generalizes across surgical datasets remains largely underexplored.

\subsection{Surgical Segmentation Datasets}
Surgical video datasets differ in which structures they annotate: instrument-only sources such as EndoVis17 (EV17) \citep{Allan2019EndoVis17} and CholecInstanceSeg (CIS) \citep{alabi2025cholecinstanceseg}, tissue-only sources such as DSAD \citep{carstens2023dresden} and SurgAI3.8k \citep{zadeh2023surgai3}, and sources covering both, such as EndoVis18 (EV18) \citep{Allan2020EndoVis18}, CholecSeg8k \citep{hong2020cholecseg8k}, and AutoLaparo \citep{wang2022autolaparo}. 
Almost all provide only per-frame semantic masks. While a few were reorganized into masklets \citep{resurgsam2}, each still covers a single procedure type.
Therefore, no existing dataset offers both the multi-procedure masklet coverage and the held-out cross-dataset splits that surgical generalization requires, a gap iSurg is built to fill.

\subsection{Memory Mechanisms and Semantic Representation Learning}
Aside from the prompted reference frame, SAM2 mainly relies on recent frames in its memory bank, which limits temporal coverage and can lead to error accumulation over extended videos.
Recent methods improve frame selection, with general approaches \citep{yang2024samurai, dam4sam, sam2long} selecting frames at inference and surgical ones \citep{masam2, surgicalsam2, resurgsam2} adapting memory to surgical scenes. 
These criteria infer identity from low-level appearance or motion cues, such as mask quality, tracking confidence, or motion continuity, which become unreliable when a distractor closely resembles the target, so the distractor may be admitted as the tracked target.
These methods also operate purely at inference, without learning long-range temporal dependencies during training.

Semantic representation learning, by contrast, has been studied mainly at the image level: vision-language approaches align visual features with category embeddings for open-vocabulary segmentation \citep{radford2021learning, lilanguage}, and prototype-based methods maintain class-level representations for few-shot segmentation \citep{wang2019panet, tian2020prior}. These inject semantic discrimination into static images, but how an object-level semantic can guide temporal memory in video tracking remains open, which is the gap SurgSLOT attempts to address.

\begin{figure}
    \centering
    \includegraphics[width=\linewidth]{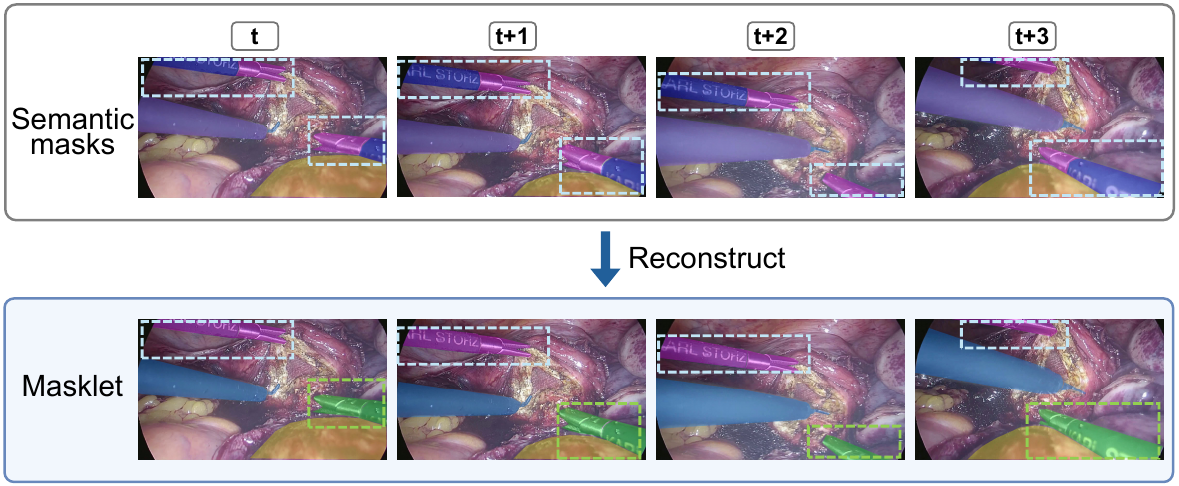}
    \caption{Reconstruction of per-frame semantic masks into temporally consistent masklets, exemplified on AutoLaparo. Part-level annotations are merged into whole-object masks, and the two same-class forceps, indistinguishable under semantic masks, receive distinct IDs that persist across frames.}
    \label{fig:recon}
\end{figure}

\section{iSurg Benchmark}
\subsection{Design Motivation}
\label{sec:motivation}
Achieving generalizable cross-dataset surgical segmentation requires a data foundation that no existing surgical dataset provides. As introduced in Sec.~\ref{sec:introduction}, the masklet annotations \citep{sam2} that iVOS needs are scarce in surgery: only EV17 and EV18 have been reorganized into masklet form \citep{resurgsam2}, while the rest provide per-frame semantic masks alone \citep{hong2020cholecseg8k, wang2022autolaparo, carstens2023dresden}, and some, such as GraSP \citep{ayobi2023matis}, are even sparse in time ($1/30$\,FPS). 
More fundamentally, these datasets, including the two reorganized into masklets, are each confined to a single procedure type, so a cross-procedure generalization model can hardly be developed based on any of them alone.

We therefore curate iSurg, which unifies $16$ surgical datasets into temporally consistent masklets across six procedure types. iSurg provides (i) \textit{masklet annotations} that support both training and propagation-based evaluation;
(ii) \textit{procedural diversity}, so that cross-procedure generalization can be both learned and assessed; and (iii) \textit{generalization-oriented splits}, in which every test set is strictly held out from training, with the split design detailed in Sec.~\ref{sec:protocol}. Together, these properties make iSurg a foundation for developing and evaluating a generalizable surgical segment-anything model.

We build these masklets along two paths over two batches of source data: a first batch with dense per-frame annotations that we reconstruct into masklets (Sec.~\ref{sec:recon}), and a second batch that is sparsely annotated or unannotated, from which a self-evolving data engine produces masklets at scale (Sec.~\ref{sec:engine}).

\subsection{Masklet Reconstruction}
\label{sec:recon}

\begin{figure*}[!tp]
\centering
    \includegraphics[width=\linewidth]{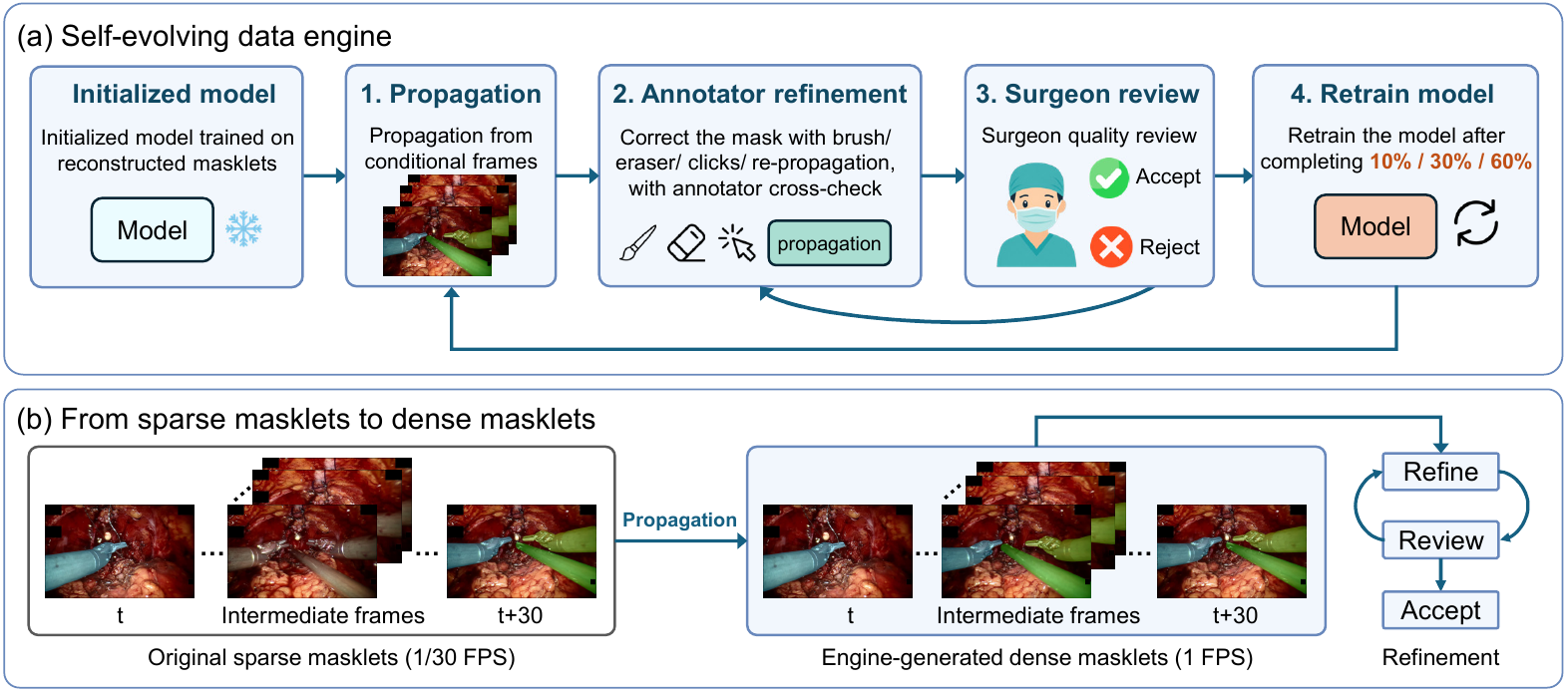}
    \caption{Self-evolving data engine for scalable masklet annotation. 
    (a) Starting from a model initialized on reconstructed masklets, each round propagates, refines, reviews, and retrains on the accepted masklets as annotation reaches $10\%$, $30\%$, and $60\%$ of the frames, so manual correction steadily decreases. 
    (b) For a temporally sparse source like GraSP, propagation densifies $1/30$\,FPS masklets to $1$\,FPS under the same refinement and review.}
    \label{fig:engine}
\end{figure*}

Reconstruction applies to datasets that already provide dense per-frame annotations.
It is necessary because these per-frame semantic masks label each pixel only by category, so they can neither separate two co-occurring objects of the same class in one frame, nor link the same object across frames. 
We reconstruct them through three steps: screening, identity assignment, and refinement and review.

\noindent\textbf{Screening.}
As the first step, we screen the original per-frame annotations and discard sequences whose masks are too noisy in boundary or temporal consistency to be reliably corrected.
Three trained medical students serve as our annotators throughout. At this screening step, all three independently inspect the original per-frame masks of each sequence frame by frame, judging temporal consistency and boundary adherence, and a sequence is discarded only if all three agree. 
This primarily affects CholecSeg8k \citep{hong2020cholecseg8k}, whose semi-automatic (Watershed) labels are often internally inconsistent and do not align with the intended structures. 
Correcting and reconstructing such sequences into usable masklets would require near-complete expert relabeling rather than refining existing reliable annotations, so we discard them instead.

\noindent\textbf{Identity Assignment.} 
Then we convert the surviving masks into masklets: because iVOS tracks a specific target rather than a category, we separate co-occurring instances of the same class and assign each a unique ID that stays fixed throughout its appearance (Fig.~\ref{fig:recon}). 
Since the vision-language supervision in Sec.~\ref{sec:tsl_vlc} reads each instrument's category name to obtain its text embedding, the same instrument is required to share one name across datasets. 
We therefore standardize instrument category names following clinical guidelines \citep{rutherford2011differentiating}, unifying the inconsistent naming across source datasets into a single label space.

\noindent\textbf{Refinement and Review.}
Screening keeps each retained masklet but does not edit it, so boundaries may still be imprecise, and part-level annotations may need merging into whole objects. 
Each masklet is therefore first refined by one annotator, who corrects the mask through iVOS-style prompting and falls back to manual pixel-level editing where prompting fails. 
Then, a second annotator cross-checks the result and flags any disagreement for the annotator to revise. 
Finally, a senior consultant surgeon reviews every masklet, approving it when the target identity remains temporally consistent and the segmentation mask is accurate, and otherwise returning it to the annotators for further correction until approval. 
Throughout this process, existing masks are only refined or removed; no new objects are introduced.

\subsection{Self-Evolving Data Engine}
\label{sec:engine}
For those datasets with sparse annotations or even without annotation, densifying annotations and scaling masklet coverage entirely by hand is prohibitively expensive at the scale cross-dataset training requires. Inspired by the model-in-the-loop data engine of SAM \citep{kirillov2023sam, sam2}, we build a self-evolving data engine that amplifies limited manual annotation into large-scale masklets: a model proposes masklets and is then retrained on the masklets it helps create, growing stronger each round so that human effort shifts progressively from labeling toward verification (Fig.~\ref{fig:engine}(a)). The engine is initialized by fine-tuning a segmentation foundation model (SAM2) on the reconstructed masklets of Sec.~\ref{sec:recon}, which adapts it to the surgical domain before any propagation.

Concretely, the engine generates masklets by propagation: a few human-labeled \textit{conditional frames} per video are propagated by the current model to all remaining frames, producing an initial masklet for every frame. 
Because each frame is initialized from a nearby labeled frame, these proposals carry only minor, locally correctable errors, and are already instance-level with identities inherited from the conditional frames. 
Each proposal then enters the same refinement-and-review as the reconstructed data (Sec.~\ref{sec:recon}): annotators refine and cross-check it, falling back to manual editing where its quality is insufficient, before the senior consultant surgeon reviews it.

We apply the data engine to two kinds of sources at scale.
For a temporally sparse source, the existing labels serve directly as conditional frames. 
This is the case for GraSP \citep{ayobi2023matis, ayobi2024pixelwise, valderrama2020tapir}, whose $1/30$\,FPS labels are propagated to a dense $1$\,FPS (Fig.~\ref{fig:engine}(b)).
For an unannotated source, a few key frames where the target clearly appears are manually labeled as conditional frames. 
This is exemplified by NUH-Hyst, an in-house hysterectomy set collected at National University Hospital, Singapore, with $4$ videos from $3$ patients, each up to $30$ minutes.

The engine runs in successive rounds from the initialized model. 
As the completed annotation reaches $10\%$, $30\%$, and $60\%$ of the frames, the model is retrained on the approved masklets, so each new round is proposed by a stronger model and needs progressively less correction.
By handling propagation, the engine confines manual effort to local correction. 
To quantify this efficiency gain, the same annotator labels $200$ instrument masks both manually and with engine assistance. Manual labeling reaches $3$ masks per minute, while engine assistance reaches $26$ per minute, a nearly $9\times$ speedup.

Throughout, every frame undergoes this same human refinement and review, so scaling up does not compromise quality.
This is particularly important for the test sources.
Among them, only the in-house NUH-Hyst is produced by the data engine, while the others are reconstructed from existing human annotations. 
For NUH-Hyst, the conditional frames are first labeled by hand on key frames where the target clearly appears, so a human specifies the targets before any propagation. The engine then drafts the remaining frames, and each draft is corrected by annotators and approved by the senior consultant surgeon before it enters the ground truth, so no test label is accepted on the basis of a model prediction alone.

The data-engine sources contribute over $120$k labeled frames and $300$k object masks at a fraction of the manual cost. We integrate this engine into a dedicated annotation system, and the supplementary video demonstrates how it supports efficient large-scale annotation. We believe this tool can contribute to the broader community by facilitating large-scale dataset creation and annotation.

\begin{table*}[!tp]
\centering
\caption{Dataset composition of the iSurg benchmark, organized by data split and surgical procedure.}
\label{tab:iSurg}
\footnotesize
\setlength{\tabcolsep}{3pt}
\begin{threeparttable}
\begin{tabular}{llrrrrrrrrr}
\toprule[1pt]
\multirow{2}{*}{Source Dataset} & \multirow{2}{*}{Procedure} & \multirow{2}{*}{Video} & \multirow{2}{*}{Frame} & \multirow{2}{*}{\makecell{Instrument\\Mask}} & \multirow{2}{*}{\makecell{Tissue\\Mask}} & \multirow{2}{*}{\makecell{Total\\Mask}} & \multirow{2}{*}{Masklet} & \multirow{2}{*}{\makecell{Duration\\(min)}} & \multirow{2}{*}{\makecell{Average\\Duration (s)}} & \multirow{2}{*}{\makecell{Label\\FPS}} \\
& & & & & & & & & & \\
\midrule
\rowcolor{gray!11}
\multicolumn{11}{c}{\textit{Training Split}}\\
Endoscapes~\citep{murali2023endoscapes} & Chole. & Img. & 468 & 747 & 1461 & 2208 & -- & -- & -- & -- \\
CholecSeg8k~\citep{hong2020cholecseg8k} & Chole. & 45 & 4338 & 6242 & 21136 & 27378 & 300 & 3 & 4 & 25 \\
CIS~\citep{alabi2025cholecinstanceseg} & Chole. & 10 & 19029 & 31416 & -- & 31416 & 77 & 317 & 1903 & 1 \\
BKAI-IGH~\citep{lan2021p} & Colon. & Img. & 1000 & -- & 1176 & 1176 & -- & -- & -- & -- \\
Kvasir-SEG~\citep{jha2019kvasir} & Colon. & Img. & 1000 & -- & 1064 & 1064 & -- & -- & -- & -- \\
ClinicDB~\citep{bernal2015wm} & Colon. & 29 & 612 & -- & 647 & 647 & 32 & 10 & 21 & 1 \\
AutoLaparo~\citep{wang2022autolaparo} & Gyn. & 300 & 1800 & 2906 & 1057 & 3963 & 738 & 30 & 6 & 1 \\
DSAD-I~\citep{carstens2023dresden} & Rect. & Img. & 10235 & -- & 10921 & 10921 & -- & -- & -- & -- \\
DSAD-V~\citep{carstens2023dresden} & Rect. & 76 & 4390 & -- & 4248 & 4248 & 91 & 73 & 58 & 1 \\
GraSP~\citep{ayobi2023matis} & Prost. & 176 & 113710 & 290926 & -- & 290926 & 993 & 1895 & 646 & 1 \\
\midrule
\textbf{Subtotal} & \textbf{-} & \textbf{636} & \textbf{156582} & \textbf{332237} & \textbf{41710} & \textbf{373947} & \textbf{2231} & \textbf{2328} & -- & -- \\
\midrule
\rowcolor{gray!11}
\multicolumn{11}{c}{\textit{Test Split}}\\
PolypGen~\citep{ali2023multi} & Colon. & 21 & 2037 & -- & 1760 & 1760 & 23 & 6 & 16 & 6 \\
Hyst-YT~\citep{fang2025spatio} & Gyn. & 6 & 1973 & 4113 & -- & 4113 & 19 & 33 & 329 & 1 \\
NUH-Hyst (in house) & Gyn. & 4 & 7200 & 15783 & -- & 15783 & 20 & 120 & 1800 & 1 \\
SurgAI3.8k~\citep{zadeh2023surgai3} & Gyn. & 51 & 3817 & -- & 3817 & 3817 & 51 & 64 & 75 & 1 \\
RARP50~\citep{psychogyios2023sar} & Prost. & 10 & 3252 & 10503 & -- & 10503 & 74 & 54 & 325 & 1 \\
EV17~\citep{Allan2019EndoVis17} & Neph. & 3 & 900 & 2265 & -- & 2265 & 10 & 15 & 300 & 1 \\
EV18~\citep{Allan2020EndoVis18} & Neph. & 4 & 596 & 1384 & 807 & 2191 & 22 & 10 & 149 & 1 \\
\midrule
\textbf{Subtotal} & \textbf{-} & \textbf{99} & \textbf{19775} & \textbf{34048} & \textbf{6384} & \textbf{40432} & \textbf{219} & \textbf{302} & -- & -- \\
\midrule
\midrule
\textbf{All} & \textbf{-} & \textbf{735} & \textbf{176357} & \textbf{366285} & \textbf{48094} & \textbf{414379} & \textbf{2450} & \textbf{2630} & -- & -- \\
\bottomrule[1pt]
\end{tabular}
\begin{tablenotes}
\footnotesize
\item Duration denotes the total video duration, and Average Duration denotes the mean duration per video. 
\item Image-only datasets are marked as ``Img.'' and excluded from the video count and duration statistics.
\item ``--'' in mask columns indicates no masks of that category exist; elsewhere it denotes a non-applicable entry.
\end{tablenotes}
\end{threeparttable}
\end{table*}

\subsection{Dataset Composition and Statistics}
\label{sec:composition}
Built through the reconstruction of public datasets (Sec.~\ref{sec:recon}) and the self-evolving data engine (Sec.~\ref{sec:engine}), 
we establish the iSurg benchmark, which consolidates $16$ surgical datasets ($15$ public and one in-house) into over $170$k frames, $410$k object masks, and $2.4$k masklets across $43$ hours ($2,630$ minutes) of video.
It spans six procedure types: cholecystectomy (Chole.), colonoscopy (Colon.), gynecology (Gyn.), rectal resection (Rect.), prostatectomy (Prost.), and nephrectomy (Neph.), with per-dataset composition, statistics, and source references detailed in Table~\ref{tab:iSurg}.
The masks span $45$ categories ($20$ instruments and $25$ tissues shown in Fig.~\ref{fig:pie}(b)), with the complete per-category breakdown provided in the supplementary material (Fig. S1).

\noindent\textbf{Data Splits and Evaluation Protocol.}
\label{sec:protocol}
iSurg is split to measure generalization to unseen sources rather than in-distribution accuracy. \textbf{All test subsets come from data sources entirely separate from the training split}, so every evaluation is inherently cross-dataset. 
On the training side, we employ a mixed image--video strategy following SAM2 to maximize data utilization. We split DSAD into image and video subsets (DSAD-I, DSAD-V): only its temporally coherent clips are reconstructed into masklets for DSAD-V, while the rest, where reliable cross-frame correspondence is hard to establish, retain DSAD's original frame-level labels as DSAD-I. CIS and GraSP supply the long videos for training long-term tracking.
For test datasets with an established split, we adopt it, and otherwise evaluate on all available data.

\noindent\textbf{Zero-Shot Procedure Type.}
EV17 and EV18 correspond to nephrectomy, a procedure type entirely absent from training, enabling zero-shot generalization to an unseen procedure type. EV18 is further divided into EV18-I and EV18-T to evaluate instrument and tissue segmentation separately. 

\noindent\textbf{Zero-Shot Category.}
By cross-checking every test category against the training label set, we identify several object categories that never appear during training: \textit{ultrasound probe}, \textit{grasping retractor}, \textit{vessel sealer}, \textit{suturing needle}, \textit{catheter}, \textit{spoon forceps}, \textit{kidney parenchyma}, and \textit{covered kidney}. 
Because these categories are verifiably absent from the training label set, the performance improvements on them indicate generalization to unseen categories.

\noindent\textbf{Long-Term Tracking.}
Beyond scale, iSurg is distinguished by the long duration of its masklets, which better captures the real-world challenges of surgical video analysis. Its videos extend up to $30$\,min, far beyond general VOS benchmarks such as SA-V \citep{sam2} ($14$\,s), MOSEv2 \citep{ding2025mosev2} ($19$\,s), and LVOS \citep{hong2023lvos} ($95$\,s). 
Over such long horizons, targets repeatedly leave and re-enter the field of view, so reliable tracking demands both long-term memory to bridge extended absences and semantic re-identification to recover the correct target on reappearance.
To evaluate this explicitly, we designate the long-duration test subsets ($\geq 300$\,s) for long-term assessment: EV17 ($300$\,s), RARP50 ($325$\,s), Hyst-YT ($329$\,s), and the in-house NUH-Hyst ($30$\,min). 
NUH-Hyst is the longest of these test subsets, several times the length of the others.
Because these four subsets are also held out from training, they jointly evaluate long-term robustness and cross-dataset generalization.

\section{Methodology}
\begin{figure*}[t]
    \centering
    \includegraphics[width=\textwidth]{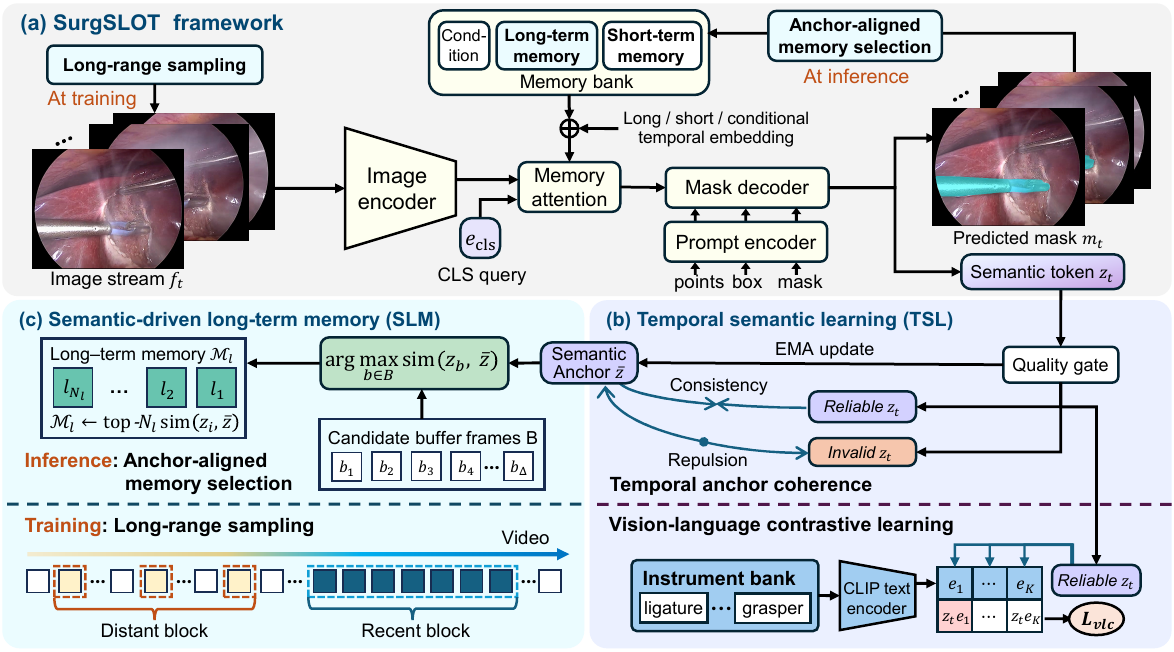}
    \caption{
    Architecture overview of SurgSLOT. 
    (a) A memory-based backbone advanced with a CLS query $e_{\textsc{cls}}$, decoded into a per-frame semantic token $z_t$. 
    (b) From the tokens $z_t$, TSL learns the semantic anchor $\bar{z}$ as the object identity: temporal anchor coherence pulls reliable $z_t$ toward $\bar{z}$ and pushes invalid ones carrying no identity evidence, while vision-language contrastive learning aligns the instrument $z_t$ with CLIP embeddings.
    (c) SLM spreads memory frames across wide temporal gaps during training, and admits the buffered frame most similar to $\bar{z}$ into long-term memory at inference.
    }
    \label{fig:surgslot}
\end{figure*}

\subsection{Overview of SurgSLOT}
\label{sec:overview}
\noindent\textbf{Problem Formulation.}
Given a surgical video $\{f_t\}_{t=1}^{T}$, $f_t\in\mathbb{R}^{3\times H\times W}$, and a first-frame prompt (point, box, or mask) $p_1$ specifying an arbitrary target instance, our goal is to produce a temporally consistent masklet $\{m_t\}_{t=1}^{T}$, $m_t\in\{0,1\}^{H\times W}$, that segments the prompted target across the video while preserving its instance identity.
The prompt specifies the target directly, so the model is not restricted to any predefined category set.
Crucially, the test video is drawn from a deployment context unseen during training, such as a new clinical center or procedure type, so a single promptable model is required to track any user-specified target from $p_1$ alone, without per-context retraining.

\noindent\textbf{SurgSLOT Architecture.}
As shown in Fig.~\ref{fig:surgslot}, SurgSLOT builds on the generic memory-based architecture of recent segmentation foundation models such as SAM2 and SAM3.
This architecture processes a video through four components: an \emph{image encoder} extracting features $F_t$; a \emph{memory bank} $\mathcal{M}_t$ storing past features and predictions; a \emph{memory attention} module conditioning $F_t$ on $\mathcal{M}_t$ to inject temporal context; 
and a \emph{mask decoder} producing $m_t$ with a predicted IoU score $u_t$ and a presence score $s_t$.
SurgSLOT reshapes the temporal modeling pathway of this architecture: it learns an object-level semantic identity and uses that identity to drive long-term memory selection.
Concretely, Temporal Semantic Learning (TSL, Sec.~\ref{sec:tsl}) learns a per-frame semantic token $z_t$ and condenses it into a slowly evolving semantic anchor $\bar{z}$ that serves as the target's stable object identity.
Then, Semantic-driven Long-term Memory (SLM, Sec.~\ref{sec:slm}) builds on this anchor to maintain a reliable long-term memory across distant frames, sustaining identity-consistent tracking over extended procedures.
The whole model is trained end-to-end under a single objective (Sec.~\ref{sec:training}).

\subsection{Temporal Semantic Learning (TSL)}
\label{sec:tsl}
TSL aims to learn a stable, object-level temporal semantic identity for each tracked target. 
This identity serves two roles: it re-identifies the target when it reappears, and it selects which past frames genuinely depict the target for long-term memory (Sec.~\ref{sec:slm}).
Such an identity requires the per-frame semantic token to have two properties: (i) the same target produces consistent tokens across frames, so its identity holds as appearance changes and after the target disappears and returns; and 
(ii) instruments of different categories produce well-separated tokens, so the target is not confused with another category of similar appearance.
However, existing representations such as SAM2's object pointer meet neither, which TSL addresses through a temporal coherence objective and a vision-language contrastive objective below.

To build this identity, TSL represents each frame's target by a semantic token $z_t$ (Fig.~\ref{fig:surgslot}(b)). 
We introduce a learnable CLS query $e_{\textsc{cls}}\in\mathbb{R}^{D}$ that produces $z_t$ through the existing memory-attention and decoder pathway without adding a separate module.
At frame $t$, $e_{\textsc{cls}}$ is concatenated with the frame features $F_t$ and passed through memory attention, where it attends to $\mathcal{M}_t$ to gather semantic context from past frames of the same target under different viewpoints, occlusions, and motion. 
It then enters the mask decoder as an additional token, processed by the same attention as the decoder's other tokens:
\begin{equation}
z_t = \big[\mathrm{Decoder}\big(\mathrm{MemAttn}([e_{\textsc{cls}},\ F_t],\ \mathcal{M}_t)\big)\big]_{\textsc{cls}},
\end{equation}
where $[\cdot]_{\textsc{cls}}$ extracts the CLS component, while $\mathrm{MemAttn}(\cdot)$ and $\mathrm{Decoder}(\cdot)$ are the base memory attention and mask decoder, respectively.
Because $z_t$ is shaped by the same attention that produces the mask, it captures the target's semantics at frame $t$ with negligible overhead.

\subsubsection{Temporal Anchor Coherence}
\label{sec:tsl_proto}
The semantic token aggregates temporal context, but this alone does not make it a stable identity: nothing constrains the semantic tokens of one tracked object to stay close across frames, so the token can drift as the target's appearance changes. 
We therefore introduce a label-free coherence objective that clusters a target's semantic tokens around a shared anchor. 
Pulling every frame's token toward one shared anchor drives it to factor out transient appearance and retain what stays common across frames. 
This invariant semantics is what the anchor carries as the object identity.
A complementary term pushes away frames that carry no valid identity evidence, so they do not contaminate the anchor.
Because the objective uses no labels, it ties this identity to no dataset-specific taxonomy, so it covers instruments and tissues alike and transfers across procedure types.

\noindent\textbf{Semantic Anchor.}
During training, for each tracked object within a clip, we maintain a slowly evolving semantic anchor $\bar{z} \in \mathbb{R}^{D}$.
It is computed as an exponential moving average (EMA) of the semantic token $z_t$ over frames where the target is reliably localized. The slow update lets it track the target's gradual appearance change while staying robust to the fluctuation of any single frame.
Let $g_t = \mathrm{IoU}(m_t, y_t)$ denote the training-only localization quality between the predicted mask $m_t$ and the ground-truth mask $y_t$, with $g_t=0$ for target-absent frames.
To prevent failed predictions from contaminating the anchor, we update it only on reliable frames, i.e., those whose $g_t$ exceeds a quality gate $\gamma_g$:
\begin{equation}
\bar{z} \leftarrow
\begin{cases}
\mathrm{norm}\big(\eta\, \bar{z} + (1-\eta)\, z_t\big), & \text{if } g_t > \gamma_g \\
\bar{z}, & \text{otherwise},
\end{cases}
\end{equation}
where $\eta$ is the EMA decay and $\mathrm{norm}(\cdot)$ denotes $\ell_2$ normalization.
The anchor is initialized from the first qualifying $z_t$ in the clip, so it accumulates only reliable evidence and is not contaminated by failed predictions.

\noindent\textbf{Consistency Loss.}
We use a margin-based loss to pull $z_t$ toward $\bar{z}$ on the reliable frames that update the anchor ($g_t > \gamma_g$):
\begin{equation}
L_{con} = \max\!\Big(0,\ \mu_{con} - \operatorname{sim}(z_t,\, \bar{z})\Big),
\end{equation}
where $\operatorname{sim}(\cdot,\cdot)$ is cosine similarity and $\mu_{con}$ is a similarity margin. Once the similarity reaches the margin $\mu_{con}$, the loss no longer penalizes the token, so $z_t$ is pulled toward the shared identity while still allowing the small appearance changes a target naturally undergoes across frames.

\noindent\textbf{Repulsion Loss.}
A frame carries no valid identity evidence when $g_t = 0$, either because the target is absent and correctly predicted empty, or because a present target is mislocalized onto the wrong region. 
In both cases, $z_t$ should not lie near the anchor, so we use a hinge to push it away from $\bar{z}$:
\begin{equation}
L_{rep} = \max\!\Big(0,\ \operatorname{sim}(z_t,\, \bar{z})\Big).
\end{equation}
Frames with intermediate quality ($0 < g_t \le \gamma_g$) are neither pulled nor pushed.

Both losses are computed with $\bar{z}$ fixed before its update at frame $t$. If the anchor has not yet been initialized, the anchor-based terms are skipped until the first reliable token initializes it. At inference, the same online anchor is maintained and reused for long-term memory selection (Sec.~\ref{sec:slm}).

\subsubsection{Vision-Language Contrastive Learning}
\label{sec:tsl_vlc}
Vision-language contrastive learning separates instrument tokens by category-level semantics, so the tracker does not confuse instruments from different categories that share a similar appearance.
We apply it to instruments only: instrument categories carry stable, transferable category-level semantics across procedures, whereas tissues vary too much across patients and surgical stages to carry such semantics.
Tissues are therefore left to the label-free coherence objective.

We obtain a base embedding for each training instrument category by feeding its name into a frozen CLIP text encoder~\citep{radford2021learning}, then pass it through a lightweight learnable projection $\phi$ to obtain the category embedding $e_k = \phi(\mathrm{CLIP}(\text{name}_k))$, yielding a pool $\{e_k\}_{k=1}^{K}$ over the $K$ training instrument categories. 
Unlike a one-hot label space, the frozen CLIP encoder places related instruments close together, providing a semantic prior. The projection $\phi$ adapts this prior knowledge to the surgical domain while preserving its structure. Because the prior carries over to instruments outside the training set, the learned discrimination extends to unseen categories, as demonstrated in Sec.~\ref{sec:zero_shot_category}.
The loss aligns $z_t$ with its positive category embedding $e_{\text{pos}}$ while pushing it away from the rest:
\begin{equation}
L_{vlc} =
\begin{cases}
-\log\dfrac{\exp(\operatorname{sim}(z_t,\, e_{\text{pos}})/\tau)}{\sum_{k=1}^{K} \exp(\operatorname{sim}(z_t,\, e_k)/\tau)}, & \text{if } g_t > \gamma_g \\[6pt]
0, & \text{otherwise},
\end{cases}
\end{equation}
where $\tau$ is a temperature \citep{liang2023open}. 
Like $L_{con}$, $L_{vlc}$ is applied only on reliable frames ($g_t > \gamma_g$), so all TSL objectives share the same ground-truth quality gate during training.
At inference, the CLIP text encoder is discarded, while $z_t$ is still computed at every frame and consumed by SLM for memory selection (Sec.~\ref{sec:slm}).

\subsection{Semantic-driven Long-term Memory (SLM)}
\label{sec:slm}
Tracking a target across a long procedure requires long-term memory, which retains earlier frames so the target can be recovered when it reappears~\citep{sam2long}. Making this memory work raises two difficulties. First, the retained frames are temporally distant from the current one, so using them means matching across long-range temporal gaps, which standard short-clip training cannot learn. Second, a frame should enter memory only if it genuinely depicts the target, yet mask quality reflects only how clean a mask is, not whether it lies on the target or a distractor.

SLM builds this capability in two stages (Fig.~\ref{fig:surgslot}(c)). 
During training, a long-range sampling strategy (Sec.~\ref{sec:slm_train}) spreads frames across long-range temporal gaps, so the memory attention learns to match across them.
At inference, an anchor-aligned selection strategy (Sec.~\ref{sec:slm_inf}) fills the memory online with frames whose semantic token stays closest to the anchor, admitting them by target identity instead of mask quality.

\subsubsection{Memory Bank Structure}
\label{sec:slm_struct}
SLM organizes the memory bank $\mathcal{M}_t$ into three components.
A \textit{conditional memory} $\mathcal{M}_c$ permanently stores the first frame $f_1$ as a fixed reference to the user's intent.
A \textit{short-term memory} $\mathcal{M}_s$ holds up to the most recent $N_s$ frames as a first-in, first-out (FIFO) queue, preserving local temporal continuity.
A \textit{long-term memory} $\mathcal{M}_l$ of capacity $N_l$ retains reliable frames from much earlier in the video.

Beyond what each component stores, the three components differ in how their temporal position is encoded. 
The conditional and short-term components keep the standard temporal embeddings, since both have a well-defined recency order relative to the current frame. 
The long-term frames are instead separated by long-range temporal gaps, where the large and irregular spacing makes their order less informative. 
We therefore give them a single shared temporal embedding that marks them as long-term references without distinguishing one from another. This shared design also leaves the long-term capacity adjustable at inference.

The three components further differ in how they are used and maintained. 
During mask decoding, the current frame attends to all three together, and when a frame qualifies for both the short-term and long-term memory, its short-term entry takes precedence to avoid duplicate attention. 
The conditional memory is fixed and never updated, and the short-term queue rolls over automatically by FIFO. 
Only the long-term memory requires active selection to decide which frames to admit and evict, detailed in Sec.~\ref{sec:slm_inf}.

\subsubsection{Long-range Sampling}
\label{sec:slm_train}
At training, long-range sampling builds the memory attention's ability to match across long-range temporal gaps, on the order of seconds to minutes, the range handled at inference.
It splits each training clip into a distant block and a recent block, processed in temporal order as a single stream. 
The distant block consists of $S_d$ frames sampled at random from a long span well before the recent block, so they are spread across a wide range and lie far from it in time.
The earliest of these serves as the conditional frame in $\mathcal{M}_c$, and the rest fill the long-term memory $\mathcal{M}_l$.
Each long-term frame is initialized interactively from its ground-truth annotation: point prompts are sampled from the mask when the target is present, and the frame yields an empty mask when the target is absent.
This keeps the long-term memory under the same predicted-mask distribution in training and inference.
The recent block consists of $S_r$ consecutive frames, over which the model propagates frame by frame using its own predictions, with the short-term memory $\mathcal{M}_s$ holding up to the most recent $N_s$ frames as a FIFO queue for local continuity.
As the stream advances, the memory accumulates frame by frame, and each frame attends to what is stored so far.

Each clip is sampled in either this long-range mode or the standard adjacent-frame mode at a fixed ratio, so training preserves short-range accuracy while enhancing long-range matching.

\subsubsection{Anchor-aligned Memory Selection}
\label{sec:slm_inf}
At inference, the whole video is tracked as one continuous sequence, so we maintain the anchor $\bar{z}$ online from the first frame onward. 
Since the ground-truth IoU is unavailable, we replace the quality gate used at training with a confidence gate $q_t > \gamma_q$, where $q_t = u_t \cdot s_t$ combines the predicted IoU $u_t$ with the object-presence score $s_t$, both produced by the decoder. 
A frame passes the gate only when its mask is both confident and on a visible target, and the anchor is updated by the EMA rule of Sec.~\ref{sec:tsl_proto} on these passing frames.

\begin{algorithm}[t]
\caption{Anchor-aligned Memory Selection at Inference}
\label{alg:slm_inf}
\SetKwInOut{Input}{Input}
\SetKwInOut{Output}{Output}
\Input{Frame stream $\{f_t\}$; semantic token $z_t$ and confidence $q_t=u_t\cdot s_t$; threshold $\gamma_q$; buffer size $\Delta$; long-term capacity $N_l$; EMA decay $\eta$ (anchor update follows Sec.~\ref{sec:tsl_proto})}
\Output{Long-term memory $\mathcal{M}_l$ maintained online}
Initialize $\mathcal{M}_l \leftarrow \emptyset$,\ buffer $B \leftarrow \emptyset$,\ anchor $\bar{z} \leftarrow \varnothing$\;
\For{each incoming frame $f_t$}{
    \eIf{$q_t > \gamma_q$}{
        Update $\bar{z}$ via EMA with $z_t$ (initialize if $\bar{z} = \varnothing$)\;
        $B \leftarrow B \cup \{f_t\}$\;
    }{
        $B \leftarrow \emptyset$ \tcp*{reset on low confidence}
    }
    \If{$|B| = \Delta$}{
        $c^* \leftarrow \arg\max_{b \in B}\ \mathrm{sim}(z_b,\ \bar{z})$\;
        $\mathcal{M}_l \leftarrow \mathcal{M}_l \cup \{c^*\}$\;
        \If{$|\mathcal{M}_l| > N_l$}{
            $\mathcal{M}_l \leftarrow \underset{m \in \mathcal{M}_l}{\operatorname{top-}N_l}\ \mathrm{sim}(z_m,\ \bar{z})$ \tcp*{keep the $N_l$ most anchor-similar frames}
        }
        $B \leftarrow \emptyset$ \tcp*{reset after admission}
    }
}
\end{algorithm}

Algorithm~\ref{alg:slm_inf} summarizes how the long-term memory $\mathcal{M}_l$ is filled. A single frame's confidence can fluctuate, so a high score on one frame alone does not mean the target is reliably tracked there. 
We therefore require continuity: a candidate buffer $B$ of size $\Delta$ collects frames while confidence stays high, and a single frame below the gate empties it and restarts collection, filtering out momentary errors. 
A full $B$ thus marks a span over which the target has been tracked reliably, and its frame closest to the anchor, $c^*$, enters $\mathcal{M}_l$. 
The memory retains only the $N_l$ frames most similar to the anchor. 
Because the anchor drifts slowly with the target's appearance, this ranking gradually drops frames that no longer match, keeping $\mathcal{M}_l$ aligned to the current identity.
Before $B$ first fills, $\mathcal{M}_l$ remains empty, and tracking relies on the conditional and short-term memory.

Selecting by anchor similarity works in two directions: a frame consistent with the anchor is admitted as a reliable reference, while a frame whose semantic token diverges from the anchor is rejected, even when its mask quality is high. 
Because the anchor is accumulated online from the test video and tied to no fixed taxonomy, this selection carries no dataset-specific prior and stays valid on procedures and categories unseen during training.

\subsection{Training Objective}
\label{sec:training} 
The full training objective combines the standard segmentation losses with the semantic learning terms of Sec.~\ref{sec:tsl}:
\begin{equation}
L_{total} = L_{base} + \lambda_{tsl}\, \big(L_{con} + L_{rep} + L_{vlc}\big),
\end{equation}
where $L_{base} = L_{dice} + L_{iou} + L_{occ} + L_{focal}$ comprises the mask overlap, IoU, object-presence, and focal losses at their default weights, and a single weight $\lambda_{tsl}$ scales the three TSL terms jointly.

\section{Experiments}
\subsection{Experimental Settings}
\label{sec:exp_settings}
\subsubsection{Prompting and Evaluation Protocol}
We evaluate on the iSurg benchmark under the cross-dataset and zero-shot protocol of Sec.~\ref{sec:protocol}. 
Visual prompts are provided only at the target's first annotated appearance, after which the model tracks the target throughout the remaining sequence, using \textbf{3-click initialization by default}.
Click placement follows SAM2's standard interactive scheme \citep{sam2,sofiiuk2022reviving}, with the first click at the mask center and subsequent clicks at the center of the largest error region. 
For mask-propagation methods such as Cutie \citep{cutie}, which require an initial mask, we use SAM2 to convert the $3$-click prompt into the first-frame mask for propagation.

For evaluation, we use the standard VOS metric $\mathcal{J}\&\mathcal{F}$ \citep{hong2023lvos}, the mean of region accuracy ($\mathcal{J}$, i.e., IoU) and boundary accuracy ($\mathcal{F}$, i.e., boundary F-score), reported as percentages.
Within each test subset, $\mathcal{J}\&\mathcal{F}$ is computed per masklet from its first appearance frame \citep{surgicalsam2} and averaged to obtain the subset-level score. The reported \textbf{Macro Average (Avg.)} throughout this paper is the average of subset-level $\mathcal{J}\&\mathcal{F}$ scores across all test subsets in iSurg. We additionally report FPS to assess inference efficiency, measured on an A6000 GPU.

\subsubsection{Implementation Details}
We build SurgSLOT on the segmentation foundation models SAM2 and SAM3 to verify that it transfers across backbones.
SurgSLOT-SAM2 uses the Hiera-B+ image encoder at $512$ resolution, while SurgSLOT-SAM3 uses the Perception Encoder at $672$, each initialized from its base model's pre-trained weights.
Models are trained for $30$ epochs with a learning rate of $1\times10^{-5}$ using mixed image–video training at a $1{:}4$ ratio, where image datasets are used for interactive image segmentation and video datasets with masklets for iVOS training. 
Video sampling uses a $1{:}1$ ratio of SLM long-range sampling and vanilla sampling. 
For TSL, the EMA anchor uses decay $\eta=0.95$, the consistency loss uses quality gate $\gamma_g=0.7$ with cosine margin $\mu_{con}=0.7$, and the contrastive loss uses the same gate with temperature $\tau=0.01$.
For SLM, each training clip uses a distant block of $S_d=3$ and a recent block of $S_r=7$ frames, with the short-term capacity fixed at $N_s=6$ following SAM2; at inference, we further set $N_l=3$, $\gamma_q=0.9$, and buffer capacity $\Delta=5$.
The loss weight for TSL is $\lambda_{tsl}=1$.
All training is conducted on four A6000 GPUs with a total batch size of $20$.

\begin{table*}[!tb]\centering
\caption{Performance comparison of iVOS methods under cross-dataset evaluation with 3-click initialization. iSurg fine-tuning generally improves performance, while SurgSLOT achieves the best overall results.}
\label{tab:three_click}
\small
\setlength{\tabcolsep}{2.5pt}
\begin{tabular}{
l
m{1.1cm}<{\centering}
m{1.1cm}<{\centering}
m{1.1cm}<{\centering}
m{1.3cm}<{\centering}
m{1.5cm}<{\centering}
m{1.3cm}<{\centering}
m{1.4cm}<{\centering}
m{1.4cm}<{\centering}
m{1cm}<{\centering}
m{1cm}<{\centering}
}
\toprule[1pt]
\multirow{2}{*}{Model}
& \multicolumn{5}{c}{Instrument}
& \multicolumn{3}{c}{Tissue}
& \multirow[c]{2}{*}[-0.3em]{\shortstack{Macro\\Avg.}}
& \multirow{2}{*}{FPS} \\
\cmidrule(l{0.5em}r{0.5em}){2-6}
\cmidrule(l{0em}){7-9}
& EV17 & EV18-I & RARP50 & Hyst-YT & NUH-Hyst
& EV18-T & SurgAI3.8k & PolypGen & & \\
\midrule
\rowcolor{gray!11}
\multicolumn{11}{c}{\textit{Vanilla Models}} \\
SAM2~\citeyearpar{sam2}          & 75.4 & 82.4 & 45.6 & 73.9 & 52.2 & 59.0 & 64.7 & 60.2 & 64.2 & 26 \\
Cutie~\citeyearpar{cutie}        & 68.7 & 79.4 & 63.3 & 82.2 & 65.2 & 67.1 & 63.4 & 54.1 & 67.9 & 53 \\
SAM2Long~\citeyearpar{sam2long}  & 72.3 & 74.4 & 42.7 & 70.1 & 46.7 & 55.7 & 66.1 & 64.2 & 61.5 & 11 \\
DAM4SAM~\citeyearpar{dam4sam}    & 72.4 & 70.7 & 39.1 & 68.3 & 47.5 & 71.0 & 65.0 & 61.5 & 61.9 & 24 \\
SAMURAI~\citeyearpar{yang2024samurai} & 68.5 & 67.7 & 38.6 & 64.1 & 39.1 & 63.9 & 64.6 & 60.8 & 58.4 & 12 \\
SurgicalSAM2~\citeyearpar{surgicalsam2} & 72.9 & 77.3 & 47.4 & 71.6 & 51.0 & 59.2 & 63.8 & 60.7 & 63.0 & 27 \\
MA-SAM2~\citeyearpar{masam2}     & 69.2 & 75.3 & 43.7 & 72.6 & 44.9 & 56.8 & 63.4 & 59.4 & 60.7 & 23 \\
MedSAM2~\citeyearpar{medsam2}    & 61.4 & 60.1 & 25.1 & 62.6 & 34.7 & 7.8  & 28.4 & 60.0 & 42.5 & 80 \\
SAM3~\citeyearpar{sam3}          & 72.5 & 78.9 & 47.5 & 73.5 & 58.1 & 70.2 & 71.5 & 64.3 & 67.1 & 9 \\
\midrule
\rowcolor{gray!11}
\multicolumn{11}{c}{\textit{iSurg Fine-Tuned Models}} \\
SAM2~\citeyearpar{sam2}          & 82.0 & 79.9 & 70.0 & 83.9 & 82.8 & 66.7 & 75.3 & 66.2 & 75.9 & 69 \\
Cutie~\citeyearpar{cutie}        & 73.8 & 75.3 & 67.4 & 78.7 & 84.4 & 59.0 & 78.6 & 63.0 & 72.5 & 53 \\
SAM2Long~\citeyearpar{sam2long}  & 77.7 & 78.9 & 60.8 & 78.6 & 69.3 & 57.4 & 76.1 & 66.3 & 70.6 & 22 \\
DAM4SAM~\citeyearpar{dam4sam}    & 76.3 & 78.1 & 57.1 & 73.9 & 66.0 & 72.0 & 74.7 & 65.2 & 70.4 & 63 \\
SAMURAI~\citeyearpar{yang2024samurai} & 70.5 & 74.8 & 57.4 & 74.5 & 67.3 & 66.5 & 75.4 & 63.7 & 68.8 & 22 \\
SurgicalSAM2~\citeyearpar{surgicalsam2} & 81.9 & 80.8 & 69.2 & 81.5 & 82.5 & 67.0 & 74.9 & 66.3 & 75.5 & 71 \\
MA-SAM2~\citeyearpar{masam2}     & 79.5 & 81.7 & 70.8 & 82.6 & 80.1 & 66.6 & 75.3 & 66.1 & 75.3 & 58 \\
SAM3~\citeyearpar{sam3}          & 82.3 & 84.3 & 72.6 & 83.1 & 80.4 & 72.3 & 78.4 & 66.2 & 77.5 & 20 \\
\rowcolor{blue!11}
SurgSLOT-SAM2 & 86.6 & 85.3 & 75.8 & \textbf{90.3} & 87.6 & 74.8 & 80.5 & 67.3 & 81.0 & 68 \\
\rowcolor{blue!11}
SurgSLOT-SAM3 & \textbf{87.6} & \textbf{90.9} & \textbf{77.0} & 88.0 & \textbf{90.8} & \textbf{76.2} & \textbf{82.2} & \textbf{69.3} & \textbf{82.8} & 20 \\
\bottomrule[1pt]
\end{tabular}
\end{table*}

\subsection{Comparison with State-of-the-Art Methods}
\label{sec:sota}
We compare SurgSLOT against representative methods spanning five groups: (1) the foundation models SAM2~\citep{sam2} and SAM3~\citep{sam3}; (2) Cutie~\citep{cutie}, a general VOS method; (3) SAM2 with general training-free memory (SAM2Long~\citep{sam2long}, DAM4SAM~\citep{dam4sam}, SAMURAI~\citep{yang2024samurai}); (4) MedSAM2~\citep{medsam2} for medical imaging; and (5) the surgical models SurgicalSAM2~\citep{surgicalsam2} and MA-SAM2~\citep{masam2}. 
We evaluate all methods using their released weights. To assess the effect of iSurg adaptation, we further fine-tune all baselines except MedSAM2 on iSurg. MedSAM2 is kept in its released setting because it already adapts SAM2 to the medical domain.
SAM2-based models use the Hiera-B+ backbone at $1024$ resolution for vanilla evaluation and $512$ after fine-tuning, since vanilla SAM2 relies on $1024$, whereas after fine-tuning, $512$ reaches almost the same accuracy at far higher speed, as discussed in Sec.~\ref{sec:discussion}.
MedSAM2, SAM3, and Cutie follow their default resolutions for vanilla evaluation, while SAM3 is fine-tuned at $672$, a multiple of its $336$ pretraining resolution chosen to balance accuracy against inference speed. Given SAM3's substantially larger size ($848$M vs.\ $80.8$M for SAM2 Hiera-B+) and lower speed ($20$ vs.\ $68$ FPS), we adopt SAM2 as the primary backbone for real-time surgical use, and evaluate on SAM3 as a higher-capacity backbone to show the framework scales to larger models.

\subsubsection{Cross-dataset Generalization}
Table~\ref{tab:three_click} reports cross-dataset results over all iSurg test subsets.
Vanilla foundation models transfer poorly to the surgical domain: SAM2 reaches only $64.2$ Macro Average $\mathcal{J}\&\mathcal{F}$, and SAM3, despite its architectural and data-scale advances, attains $67.1$, indicating that general-domain progress alone does not bridge the surgical gap. 
MedSAM2 reaches only $42.5$ despite medical fine-tuning, suggesting that the modality heterogeneity in its training data hinders transfer to surgical video. 
Among the methods fine-tuned on iSurg, surgical-specific training consistently improves the Macro Average score, underscoring the value of the proposed data foundation.
SurgSLOT-SAM2 attains the best result among SAM2-based models with $81.0$ Macro Average at $68$ FPS real-time inference, improving over vanilla and fine-tuned SAM2 by $16.8$ and $5.1$ points, respectively.
The same gains carry over to SAM3, where SurgSLOT attains the overall best $82.8$ Macro Average, improving over fine-tuned SAM3 by $5.3$ points.
These consistent gains on both SAM2 and SAM3 suggest that SurgSLOT is not tied to a specific backbone, but instead provides a general approach for memory-based segmentation foundation models.
Among the fine-tuned baselines, some training-free memory methods improve over fine-tuned SAM2 on tissues but degrade on instruments (e.g., DAM4SAM gains $5.3$ on EV18-T). Tissues mostly stay in view, whereas instruments leave and re-enter it, which we analyze next.

\subsubsection{Long-term Tracking Analysis}
\label{sec:long_term}
Surgical procedures span tens of minutes, during which frequent camera motion and repeated target disappearance break the track. This challenge is most acute for instruments, which leave the view and need to be re-identified on reappearance, while tissues stay largely visible and rarely require re-identification.
We assess this instrument tracking challenge on the long-duration subsets EV17 ($300$\,s), RARP50 ($325$\,s), Hyst-YT ($329$\,s), and the in-house NUH-Hyst ($30$\,min), with per-subset $\mathcal{J}\&\mathcal{F}$ reported in Table~\ref{tab:three_click}.
In both vanilla and fine-tuned settings, training-free memory methods (SAM2Long, DAM4SAM, SAMURAI) degrade relative to SAM2 on these subsets. Their selection rules discard the target-absent frames, so during the long disappearances caused by prolonged camera motion, the memory retains only outdated references from before the target left, so it lacks a continuous reference close in time to the current frame.
With neither a record of the absence nor a continuous temporal reference, a later similar instrument may be misidentified as the target, producing persistent false positives.
MA-SAM2 alleviates this with a surgical-specific memory design. Yet it selects memory frames by mask quality and distractor prediction, which can admit a clean mask on the wrong object and propagate the error, so it does not improve over fine-tuned SAM2 on most of these subsets.
In contrast, SurgSLOT pairs long-term memory with a semantic anchor that re-identifies the target after disappearance and rejects similar distractors, improving over fine-tuned SAM2 on all four subsets. 
Averaged over all four subsets, SurgSLOT-SAM2 rises from $79.7$ to $85.1$ ($+5.4$), and on the longest NUH-Hyst improves from $82.8$ to $87.6$.

\begin{figure}
    \centering
    \includegraphics[width=\linewidth]{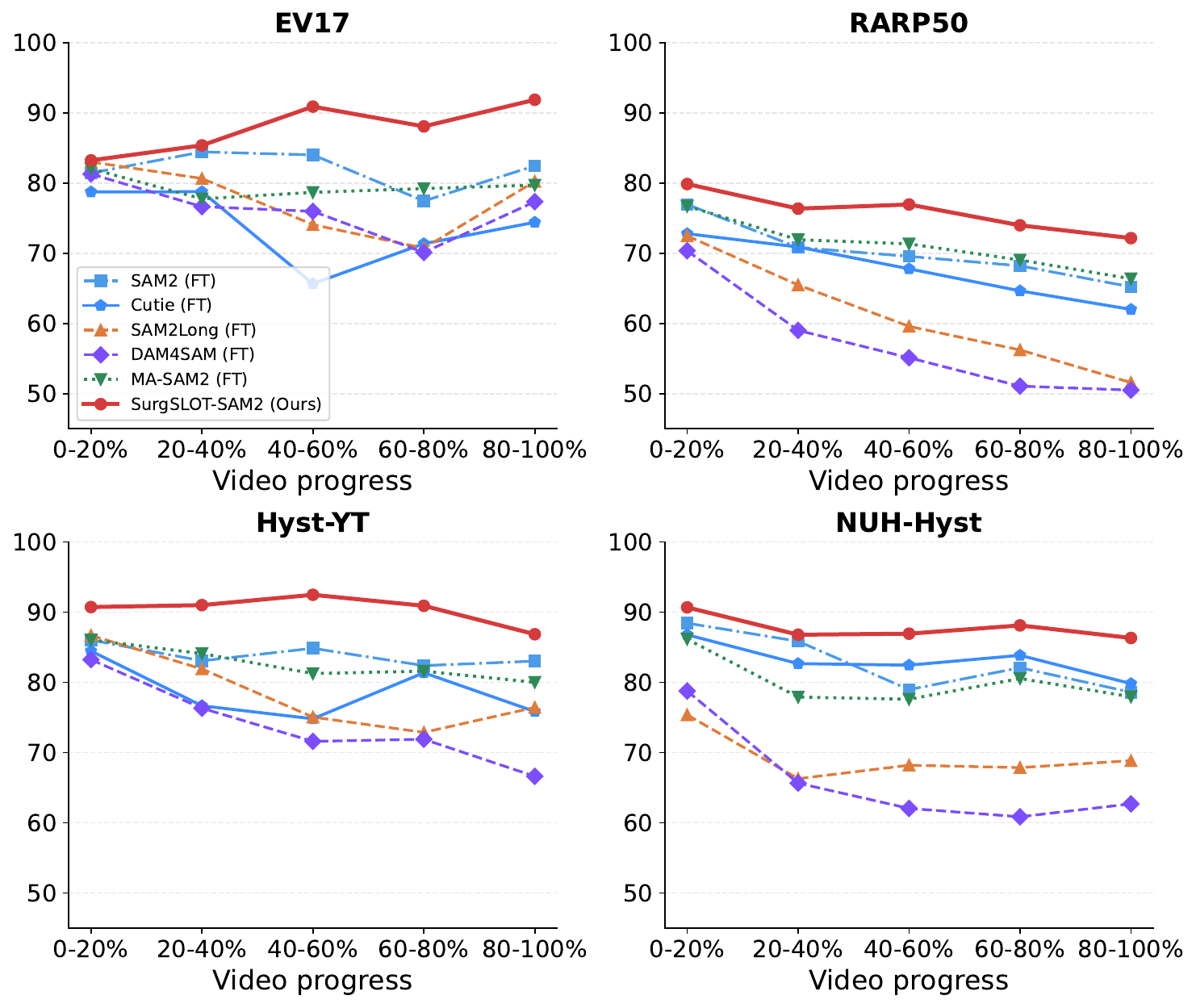}
    \caption{$\mathcal{J}$\&$\mathcal{F}$ across video progress intervals on long-duration subsets under 3-click initialization. Vanilla models are omitted due to their substantially lower overall performance (Table~\ref{tab:three_click}).}
    \label{fig:long_term_comparison}
\end{figure}

To further examine robustness over time, we track each video throughout from the single first-frame prompt, then split the completed track into five equal-length temporal segments and report the average $\mathcal{J}\&\mathcal{F}$ within each (Fig.~\ref{fig:long_term_comparison}). 
For clarity, we plot representative fine-tuned methods and omit vanilla models, given their substantially lower overall accuracy. 
Most of these baselines degrade noticeably as the video progresses, especially on the challenging RARP50 and the longest NUH-Hyst, where occlusions from blood, smoke, and instrument interactions compound over time.
In contrast, SurgSLOT-SAM2 maintains consistently high performance across all four long-duration subsets with only minor degradation, indicating that its anchor-driven long-term memory suppresses the error accumulation that compounds in extended surgical videos.

\subsubsection{Zero-shot Procedure Type Generalization}
On EV17 and EV18, held out from training as the zero-shot procedure-type setting (Sec.~\ref{sec:protocol}), both SurgSLOT variants surpass all vanilla and fine-tuned baselines on EV17, EV18-I, and EV18-T, with SurgSLOT-SAM3 reaching up to $90.9$ $\mathcal{J}\&\mathcal{F}$ on EV18-I. 
The gains hold on both instruments (EV17, EV18-I) and tissue (EV18-T), so the generalization is not limited to one structure type. Because the semantic anchor encodes a target's identity without binding it to any procedure-specific appearance, it transfers to an unseen procedure and re-identifies targets there.
These results indicate that SurgSLOT generalizes to unseen procedure types instead of memorizing procedure-specific appearance.

\begin{table*}[!tb]
\centering
\begin{threeparttable}
\caption{Performance comparison of iVOS methods under zero-shot category evaluation with 3-click initialization. Average is the unweighted mean $\mathcal{J}$\&$\mathcal{F}$ across unseen categories, and SurgSLOT achieves the best overall performance.}
\label{tab:zero_shot_category}
\footnotesize
\setlength{\tabcolsep}{4pt}
\begin{tabular}{
l
>{\centering\arraybackslash}m{1.35cm}
>{\centering\arraybackslash}m{1.35cm}
>{\centering\arraybackslash}m{1.25cm}
>{\centering\arraybackslash}m{1.25cm}
>{\centering\arraybackslash}m{1.15cm}
>{\centering\arraybackslash}m{1.30cm}
>{\centering\arraybackslash}m{1.45cm}
>{\centering\arraybackslash}m{1.35cm}
>{\centering\arraybackslash}m{1.05cm}
}
\toprule[1pt]
\multirow{3}{*}{Model}
& \multicolumn{6}{c}{Instrument}
& \multicolumn{2}{c}{Tissue}
& \multirow{3}{*}{Average} \\
\cmidrule(l{0.5em}r{0.5em}){2-7}
\cmidrule(l{0.5em}r{0.5em}){8-9}
& \shortstack{Ultrasound\\Probe}
& \shortstack{Grasping\\Retractor}
& \shortstack{Vessel\\Sealer}
& \shortstack{Suturing\\Needle}
& Catheter
& \shortstack{Spoon\\Forceps}
& \shortstack{Kidney\\Parenchyma}
& \shortstack{Covered\\Kidney}
& \\
\midrule
\rowcolor{gray!11}
\multicolumn{10}{c}{\textit{Vanilla Models}} \\
Cutie~\citeyearpar{cutie}        & 75.1 & 54.2 & 74.1 & 27.2 & 69.5 & 60.2 & 73.5 & 50.5 & 60.5 \\
SAM2~\citeyearpar{sam2}          & 59.3 & 50.7 & 88.7 & 30.4 & 71.9 & 22.7 & 54.4 & 52.0 & 53.8 \\
MA-SAM2~\citeyearpar{masam2}     & 65.6 & 16.6 & 90.0 & 27.1 & 74.0 & 7.3  & 56.0 & 50.1 & 48.3 \\
SAM3~\citeyearpar{sam3}          & 63.9 & 13.2 & 92.8 & 35.1 & 76.5 & 37.7 & 80.4 & 50.2 & 56.2 \\
\midrule
\rowcolor{gray!11}
\multicolumn{10}{c}{\textit{iSurg Fine-Tuned Models}} \\
Cutie~\citeyearpar{cutie}        & 83.8 & 77.2 & 39.4 & 26.1 & 68.8 & 78.6 & 58.7 & 52.2 & 60.6 \\
SAM2~\citeyearpar{sam2}          & 83.7 & 77.5 & 66.6 & 29.9 & 71.1 & 77.2 & 63.7 & 50.8 & 65.1 \\
MA-SAM2~\citeyearpar{masam2}     & 81.9 & 40.8 & 89.7 & 28.7 & 72.5 & 50.3 & 63.3 & 50.0 & 59.7 \\
SAM3~\citeyearpar{sam3}          & 78.0 & 77.4 & 89.3 & 32.9 & 80.4 & 47.2 & 75.2 & 52.1 & 66.6 \\
\rowcolor{blue!11}
SurgSLOT-SAM2  & 84.2 & \textbf{78.2} & 91.1 & 35.4 & 74.6 & 77.4 & 81.6 & \textbf{53.0} & 71.9 \\
\rowcolor{blue!11}
SurgSLOT-SAM3  & \textbf{90.5} & \textbf{78.2} & \textbf{93.8} & \textbf{40.2} & \textbf{87.1} & \textbf{87.3} & \textbf{84.1} & 52.5 & \textbf{76.7} \\
\bottomrule[1pt]
\end{tabular}
\end{threeparttable}
\end{table*}

\subsubsection{Zero-shot Category Generalization}
\label{sec:zero_shot_category}
Beyond unseen procedure types, we evaluate generalization to object categories that never appear during training. Eight categories are held out entirely (six instruments and two tissues listed in Table~\ref{tab:zero_shot_category}), so the successful segmentation cannot be attributed to category memorization and instead reflects the model's ability to segment and track genuinely unseen targets.
For this fine-grained per-category breakdown, we compare against representative methods spanning general VOS (Cutie), foundation backbones (SAM2, SAM3), and surgical adaptation (MA-SAM2), keeping the per-category table compact.

SurgSLOT achieves the best overall results, with SurgSLOT-SAM2 and SurgSLOT-SAM3 reaching $71.9$ and $76.7$ average $\mathcal{J}\&\mathcal{F}$, surpassing their fine-tuned baselines by $6.8$ and $10.1$ points.
The gains span both unseen instruments and unseen tissues, which TSL handles through its two complementary objectives. 
The label-free coherence objective forms an object-level identity for any tracked target without relying on a category label, so it extends to unseen tissues such as \textit{kidney parenchyma}.
On top of this, the vision-language supervision (Sec.~\ref{sec:tsl_vlc}) organizes instrument categories by transferable semantics, so the discrimination learned among seen instruments carries over to unseen ones such as \textit{ultrasound probe}, \textit{catheter}, and \textit{spoon forceps}.
The same anchor selects long-term memory by semantic consistency, so its selection carries over to these unseen instruments and tissues without modification.
Both backbones extend to these unseen instruments, with SurgSLOT-SAM3 pushing them the highest as its larger capacity better exploits the transferable instrument semantics.

Two patterns warrant discussion. 
First, fine-tuning does not uniformly help every unseen category. On \textit{vessel sealer}, vanilla SAM2 reaches $88.7$, but fine-tuning drops it to $66.6$, a negative transfer that SurgSLOT-SAM2 avoids by holding $91.1$.
Second, the \textit{suturing needle} remains difficult for all methods due to its tiny structure and frequent occlusion, with the best competing method (vanilla SAM3) reaching only $35.1$. Nevertheless, SurgSLOT-SAM3 attains the highest score of $40.2$, a $5.1$-point gain over that baseline.
Overall, the consistent advantage in unseen categories confirms that SurgSLOT generalizes not only to the unseen procedure types but also to unseen categories.

\subsubsection{Robustness to Prompt Types}
\label{sec:prompt_types}
\begin{table}[!tp]
\centering
\caption{Macro Average $\mathcal{J}\&\mathcal{F}$ across all iSurg test subsets under five prompt types.}
\label{tab:prompt_comparison}
\footnotesize
\setlength{\tabcolsep}{3pt}
\begin{tabular}{lccccc}
\toprule[1pt]
Model & 1-click & 3-click & 5-click & BBox & GT mask \\
\midrule
\rowcolor{gray!11}
\multicolumn{6}{c}{\textit{Vanilla Models}} \\
SAM2~\citeyearpar{sam2}          & 54.4 & 64.2 & 67.3 & 66.4 & 68.1 \\
Cutie~\citeyearpar{cutie}        & 59.4 & 67.9 & 69.7 & 68.2 & 71.4 \\
SAM2Long~\citeyearpar{sam2long}  & 52.8 & 61.5 & 64.2 & 63.6 & 65.2 \\
DAM4SAM~\citeyearpar{dam4sam}    & 53.6 & 61.9 & 62.8 & 63.5 & 64.3 \\
SAMURAI~\citeyearpar{yang2024samurai} & 50.0 & 58.4 & 60.4 & 59.7 & 61.5 \\
SurgicalSAM2~\citeyearpar{surgicalsam2} & 54.6 & 63.0 & 65.3 & 65.6 & 66.6 \\
MA-SAM2~\citeyearpar{masam2}     & 53.3 & 60.7 & 64.6 & 64.7 & 64.6 \\
MedSAM2~\citeyearpar{medsam2}    & 40.2 & 42.5 & 43.9 & 42.9 & 44.1 \\
SAM3~\citeyearpar{sam3}          & 56.6 & 67.1 & 68.2 & 67.3 & 68.8 \\
\midrule
\rowcolor{gray!11}
\multicolumn{6}{c}{\textit{iSurg Fine-Tuned Models}} \\
SAM2~\citeyearpar{sam2}          & 71.5 & 75.9 & 76.2 & 75.6 & 76.4 \\
Cutie~\citeyearpar{cutie}        & 69.1 & 72.5 & 75.9 & 74.8 & 76.3 \\
SAM2Long~\citeyearpar{sam2long}  & 66.9 & 70.6 & 71.8 & 71.6 & 72.6 \\
DAM4SAM~\citeyearpar{dam4sam}    & 66.0 & 70.4 & 71.2 & 68.5 & 72.4 \\
SAMURAI~\citeyearpar{yang2024samurai} & 64.5 & 68.8 & 69.6 & 68.5 & 69.6 \\
SurgicalSAM2~\citeyearpar{surgicalsam2} & 71.2 & 75.5 & 76.3 & 75.4 & 76.6 \\
MA-SAM2~\citeyearpar{masam2}     & 69.7 & 75.3 & 74.8 & 74.7 & 75.6 \\
SAM3~\citeyearpar{sam3}          & 73.3 & 77.5 & 78.2 & 77.9 & 78.3 \\
\rowcolor{blue!11}
SurgSLOT-SAM2 & 77.5 & 81.0 & 81.7 & 80.4 & 81.9 \\
\rowcolor{blue!11}
SurgSLOT-SAM3 & \textbf{78.8} & \textbf{82.8} & \textbf{83.4} & \textbf{82.9} & \textbf{83.8} \\
\bottomrule[1pt]
\end{tabular}
\end{table}

To assess sensitivity to initialization, we evaluate all methods under five prompt types: $1$-click, $3$-click, $5$-click, bounding box, and ground-truth mask, reporting the Macro Average $\mathcal{J}\&\mathcal{F}$ across all iSurg test subsets in Table~\ref{tab:prompt_comparison}. 
SurgSLOT-SAM3 attains the best result under every prompt type, with SurgSLOT-SAM2 ranking second across all settings. Notably, SurgSLOT-SAM2 with a single click ($77.5$) already exceeds fine-tuned SAM2 initialized with a ground-truth mask ($76.4$), indicating that the proposed modules substantially reduce the dependence on detailed initialization. 
The consistent ranking across all five prompt types confirms that SurgSLOT's advantage stems from its tracking design rather than from any particular prompt format.

\subsubsection{Qualitative Analysis}
\begin{figure*}
    \centering
    \includegraphics[width=\textwidth]{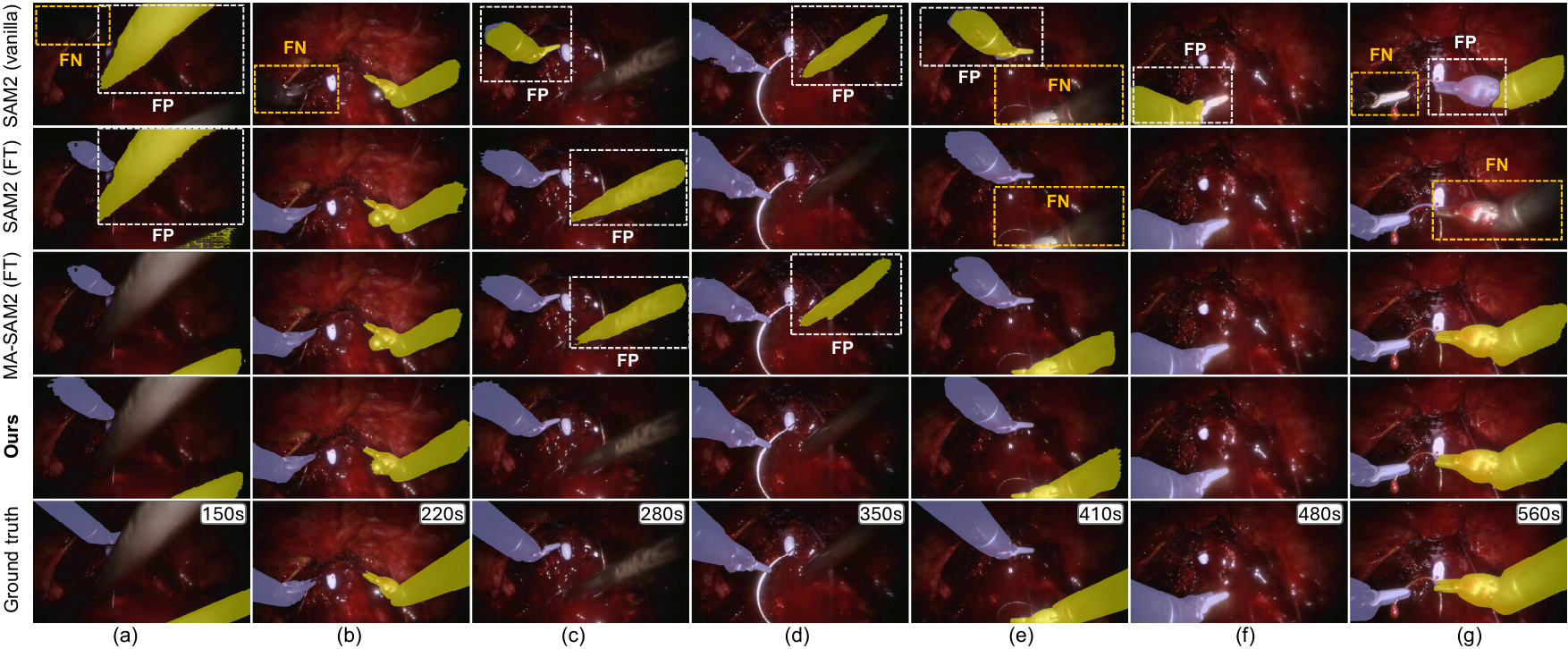}
    \caption{Qualitative comparison between SAM2 (vanilla), SAM2 (FT), MA-SAM2 (FT), and SurgSLOT-SAM2 on RARP50. FT denotes fine-tuned models. Frame indices indicate timestamps in seconds, spanning from 150s to 560s (410s duration).
    }
    \label{fig:quality_rarp}
\end{figure*}

\begin{figure}
    \centering
    \includegraphics[width=\linewidth]{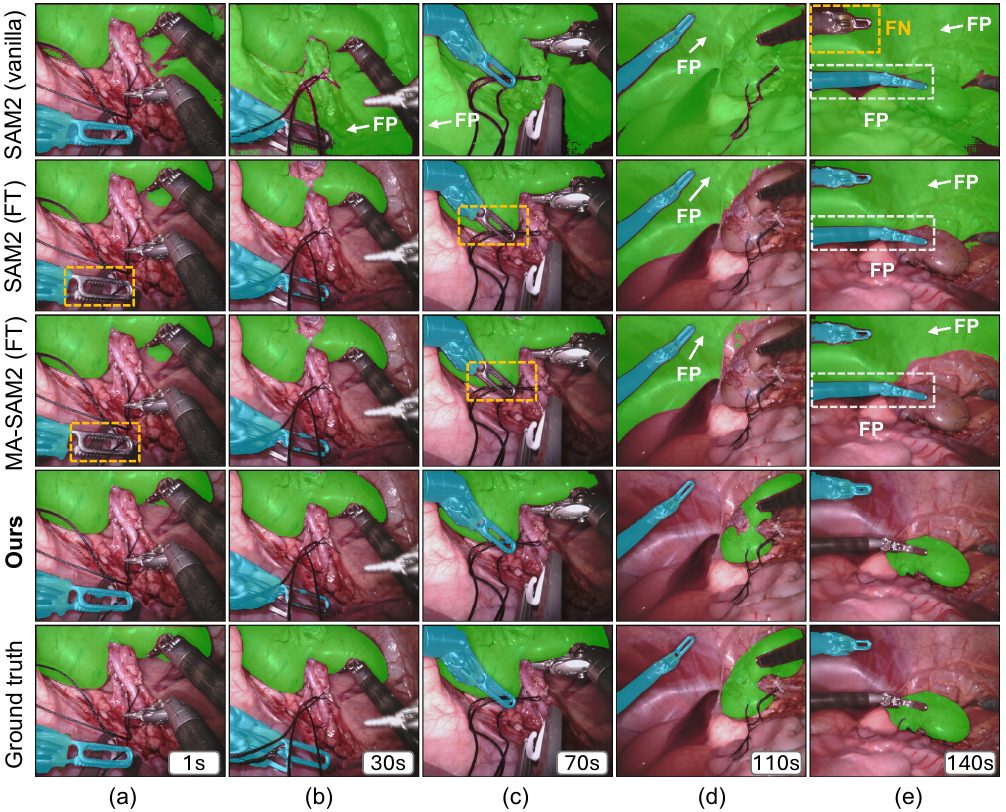}
    \caption{Qualitative comparison between SAM2 (vanilla), SAM2 (FT), MA-SAM2 (FT), and SurgSLOT-SAM2 on EV18, with 140s duration.}
    \label{fig:quality_ev18}
\end{figure}
Figs.~\ref{fig:quality_rarp} and \ref{fig:quality_ev18} compare SurgSLOT-SAM2 with vanilla SAM2, fine-tuned SAM2, and fine-tuned MA-SAM2 on RARP50 and EV18 under 3-click initialization.
Vanilla SAM2 exhibits frequent false positives (FPs) and false negatives (FNs) under challenging conditions such as blood occlusions, lighting variations, and dynamic scene changes.
Fine-tuned SAM2 and MA-SAM2 improve robustness, yet still produce FPs and incomplete segmentation at instrument edges. 
On RARP50, when the tracked instrument exits and a visually similar one appears, all baselines produce FPs (Fig.~\ref{fig:quality_rarp}(c,d)), 
whereas SurgSLOT-SAM2 leverages TSL's semantic discrimination to reject the distractor and re-identify the original target upon reappearance (Fig.~\ref{fig:quality_rarp}(e)).
On EV18, all baselines suffer from progressive tracking drift on tissues, with vanilla SAM2 losing the target earliest and fine-tuned methods drifting in later frames (Fig.~\ref{fig:quality_ev18}(c,d)). 
In contrast, SurgSLOT-SAM2 keeps the semantic token aligned to its anchor and selects long-term memory by that identity, so it tracks the tissue stably with precise boundaries across all frames and avoids the mask drift exhibited by the baselines.
Overall, SurgSLOT-SAM2 maintains accurate segmentation and target identity throughout long-duration videos, illustrating the robustness of the proposed design for long-term surgical tracking.
Full-video comparisons between SurgSLOT-SAM2 and vanilla SAM2 are provided in the supplementary video, demonstrating real-time long-term tracking at $68$ FPS across EV18, RARP50, and NUH-Hyst.

\begin{table}[!tp]\centering
\caption{Comparison on nephrectomy between in-domain iVOS specialists and iSurg models, the latter under zero-shot evaluation. Reported in $\mathcal{J}\&\mathcal{F}$ at $512$ resolution.}
\label{tab:in_domain}
\footnotesize
\setlength{\tabcolsep}{6pt}
\begin{tabular}{lcccc}
\toprule[1pt]
Model & EV17 & EV18-I & EV18-T \\
\midrule
SAM2 (vanilla)        & 63.1 & 72.2 & 46.4 \\
SAM2 (specialist)     & 75.9 & 76.7 & 74.4 \\
SAM2 (iSurg)          & 82.0 & 79.9 & 66.7 \\
\rowcolor{blue!11}
SurgSLOT-SAM2 (iSurg) & \textbf{86.6} & \textbf{85.3} & \textbf{74.8} \\
\bottomrule[1pt]
\end{tabular}
\end{table}

\subsection{Comparison with Other Segmentation Paradigms}
\label{sec:paradigm}
The preceding comparisons are confined to the iVOS paradigm.
To position SurgSLOT as a generalist within the broader landscape of surgical segmentation, we further compare it against two families of in-domain specialists, each trained on its target dataset: iVOS specialists (the per-dataset counterpart of SurgSLOT) and semantic segmentation methods (the conventional choice for surgical instrument segmentation, which seldom covers tissues).
Unlike these specialists, SurgSLOT receives no in-domain training, so these comparisons test whether one promptable generalist can match dataset-specific specialists.

\begin{table}[!tp]\centering
\caption{Instrument segmentation comparison with in-domain semantic methods on EV17 and EV18. Accuracy is reported as CIoU.}
\label{tab:vs_semantic}
\footnotesize
\setlength{\tabcolsep}{6pt}
\begin{threeparttable}
\begin{tabular}{lccc}
\toprule[1pt]
Model & EV17 & EV18-I & FPS \\
\midrule
\rowcolor{gray!11}
\multicolumn{4}{c}{\textit{In-domain semantic segmentation}} \\
ISINet~\citeyearpar{gonzalez2020in}        & 55.6 & 73.0 & 16 \\
SurgicalSAM~\citeyearpar{Yue2024surgsam}   & 69.9 & 80.3 & 2 \\
MATIS (Full$^\ast$)~\citeyearpar{ayobi2023matis}  & 71.4 & 84.3 & 15 \\
LACOSTE (B$^\dagger$)~\citeyearpar{wang2025lacoste}  & 82.3 & 86.5 & 3 \\
LACOSTE (L$^\dagger$)~\citeyearpar{wang2025lacoste}  & 83.4 & 86.8 & 2 \\
\midrule
\rowcolor{gray!11}
\multicolumn{4}{c}{\textit{Vanilla iVOS, zero-shot}} \\
SAM2~\citeyearpar{sam2}  & 77.8 & 81.8 & 26 \\
SAM3~\citeyearpar{sam3}  & 81.6 & 82.4 & 9 \\
\midrule
\rowcolor{gray!11}
\multicolumn{4}{c}{\textit{iSurg fine-tuned iVOS, zero-shot cross-procedure type}} \\
SAM2~\citeyearpar{sam2} (iSurg)        & 79.7 & 80.5 & 69 \\
MA-SAM2~\citeyearpar{masam2} (iSurg)   & 79.3 & 82.2 & 58 \\
SAM3~\citeyearpar{sam3} (iSurg)        & 81.5 & 83.8 & 20 \\
\rowcolor{blue!11}
SurgSLOT-SAM2    & \textbf{85.6} & 85.1 & 68 \\
\rowcolor{blue!11}
SurgSLOT-SAM3    & 84.8 & \textbf{88.2} & 20 \\
\bottomrule[1pt]
\end{tabular}
\begin{tablenotes}[flushleft]
\footnotesize
\item[$^\ast$] Full temporal variant of MATIS.
\item[$^\dagger$] B/L denote the Base and Large variants of LACOSTE.
\end{tablenotes}
\end{threeparttable}
\end{table}
\subsubsection{Comparison with In-domain iVOS Specialists}
\label{sec:vs_specialist}
A natural concern with any generalist model is whether sacrificing in-domain specialization is worthwhile. To answer this, we compare SurgSLOT-SAM2 against SAM2 specialists trained and evaluated on each in-domain training split, alongside vanilla SAM2 as a lower bound and SAM2 fine-tuned on iSurg as a same-data control.
The setting favors the specialists: EV17 and EV18 are nephrectomy, a procedure absent from iSurg training, so both iSurg models are evaluated zero-shot there, whereas the specialists are trained in-domain.
Table~\ref{tab:in_domain} compares these models under $3$-click initialization, reporting $\mathcal{J}\&\mathcal{F}$.
Despite this zero-shot disadvantage, a single SurgSLOT-SAM2 model surpasses the in-domain specialists on all three subsets, with sizable margins on EV17 ($+10.7$) and EV18-I ($+8.6$) and an edge on EV18-T ($+0.4$).
This in-domain advantage does not translate into higher accuracy because EV17 and EV18 are too small to fit a robust specialist, and the specialists carry no long-term memory to resist drift over long nephrectomy sequences.
Taken together, these results show that a single SurgSLOT model trained on iSurg generalizes across diverse surgical scenarios without per-dataset fine-tuning, while surpassing in-domain iVOS specialists even on a procedure type held out from its training.

\subsubsection{Comparison with Instrument Semantic Segmentation}
\label{sec:vs_semantic}

\begin{table*}[!tp]\centering
\caption{Component ablation of SurgSLOT under 3-click initialization. TSL components are added cumulatively, followed by the two components of SLM.}
\label{tab:component_ablation}
\footnotesize
\setlength{\tabcolsep}{2pt}
\begin{threeparttable}
\begin{tabular}{lccccccccccc}
\toprule[1pt]
\multirow{2}{*}{Configuration}
& \multicolumn{5}{c}{Instrument}
& \multicolumn{3}{c}{Tissue}
& \multirow[c]{2}{*}[-0.3em]{\makecell{Instrument\\Avg.}}
& \multirow[c]{2}{*}[-0.3em]{\makecell{Tissue\\Avg.}}
& \multirow[c]{2}{*}[-0.3em]{\makecell{Macro\\Avg.}} \\
\cmidrule(l{0.5em}r{0.5em}){2-6}
\cmidrule(l{0.5em}r{0.5em}){7-9}
& EV17 & EV18-I & RARP50 & Hyst-YT & NUH-Hyst
& EV18-T & SurgAI3.8k & PolypGen
& & & \\
\midrule
SAM2 (vanilla)
& 63.1 & 72.2 & 47.2 & 65.7 & 68.7
& 46.4 & 45.3 & 53.8
& 63.4 & 48.5 & 57.8 \\
\textit{+ iSurg FT}
& 82.0 & 79.9 & 70.0 & 83.9 & 82.8
& 66.7 & 75.3 & 66.2
& 79.7 & 69.4 & 75.9 \\
\rowcolor{gray!11}
\multicolumn{12}{l}{\textit{TSL component ablation}} \\
\textit{+ Consistency loss}
& 82.5 & 82.1 & 72.2 & 85.3 & 83.5
& 68.9 & 77.7 & 66.3
& 81.1 & 71.0 & 77.3 \\
\textit{+ Repulsion loss}
& 82.9 & 82.5 & 72.5 & 85.1 & 84.2
& 69.4 & 77.7 & 66.7
& 81.4 & 71.3 & 77.6 \\
\textit{+ VLC}
& 83.5 & 83.3 & 73.1 & 86.7 & 84.7
& 70.6 & 78.1 & 66.6
& 82.3 & 71.8 & 78.3 \\
\rowcolor{gray!11}
\multicolumn{12}{l}{\textit{SLM component ablation}} \\
\textit{+ Long-range sampling}$^\ast$
& 84.0 & 83.9 & 73.7 & 88.1 & 85.5
& 72.5 & 78.6 & 66.8
& 83.0 & 72.6 & 79.1 \\
\rowcolor{blue!11}
\textit{+ Anchor-aligned memory}
& \textbf{86.6} & \textbf{85.3} & \textbf{75.8} & \textbf{90.3} & \textbf{87.6}
& \textbf{74.8} & \textbf{80.5} & \textbf{67.3}
& \textbf{85.1} & \textbf{74.2} & \textbf{81.0} \\
\bottomrule[1pt]
\end{tabular}
\begin{tablenotes}[flushleft]
\footnotesize
\item[$^\ast$] Use vanilla recent-frame memory before long-term selection.
\end{tablenotes}
\end{threeparttable}
\end{table*}

A large body of surgical instrument segmentation work follows the traditional semantic segmentation paradigm and reports Challenge IoU (CIoU) on EV17 and EV18. 
We therefore compare SurgSLOT against representative in-domain semantic segmentation methods, including ISINet~\citep{gonzalez2020in}, SurgicalSAM~\citep{Yue2024surgsam}, MATIS~\citep{ayobi2023matis}, and LACOSTE~\citep{wang2025lacoste}, as summarized in Table~\ref{tab:vs_semantic}, together with iVOS baselines with SAM2~\citep{sam2}, SAM3~\citep{sam3}, and MA-SAM2~\citep{masam2}. 

These two paradigms address different problems and are not directly comparable. Semantic segmentation assigns every pixel a label from a fixed set of categories, so it handles only those predefined classes, whereas SurgSLOT segments and tracks any prompt-specified instance, extending to categories never seen in training. To place them under a common metric, we evaluate both with CIoU, the per-class measure used by these semantic methods, aggregating SurgSLOT's per-instance masklets into per-frame semantic label maps.

The setting favors the semantic baselines: they are trained \textbf{in-domain} on nephrectomy, whereas SurgSLOT runs \textbf{zero-shot}, never trained on any nephrectomy data. This comparison therefore tests whether a single zero-shot generalist can match procedure-specific in-domain specialists. We follow the standard 4-fold protocol on EV17 with identical fold partitions, and measure FPS end-to-end on the same A6000 GPU at default settings. Because this evaluation differs from Table~\ref{tab:three_click} in both data partition and metric, the EV17 scores are not comparable across the two tables.

Despite the zero-shot disadvantage, SurgSLOT remains highly competitive. On EV17, SurgSLOT-SAM2 surpasses the strongest in-domain baseline LACOSTE (L) by $2.2$ points ($85.6$ vs.\ $83.4$). On EV18-I, SurgSLOT-SAM2 trails LACOSTE (L) ($85.1$ vs.\ $86.8$), as expected given the latter's in-domain training, but SurgSLOT-SAM3 reverses this and outperforms all in-domain methods ($88.2$). SurgSLOT is also far more efficient: LACOSTE runs at only $2$--$3$ FPS, while SurgSLOT-SAM2 reaches $68$ FPS. 
Relying only on a first-frame prompt, a single zero-shot generalist thus remains competitive with in-domain semantic specialists in accuracy while running far faster.

\subsection{Ablation Study and Detailed Analysis}

\subsubsection{Component Ablation}
We perform ablation studies to evaluate each component, as shown in Table~\ref{tab:component_ablation}, reporting all iSurg test subsets grouped into instrument and tissue.
Instrument Avg.\ and Tissue Avg.\ are the unweighted means of $\mathcal{J}\&\mathcal{F}$ over the five instrument subsets and three tissue subsets, and Macro Avg.\ averages over all eight. 
All rows are evaluated at $512$ resolution, so the vanilla SAM2 baseline here ($57.8$) differs from its $1024$-resolution counterpart ($64.2$). 
Components are added cumulatively in dependency order: TSL first learns the semantic anchor, then SLM uses it to manage long-term memory, with TSL and SLM each broken down into their sub-components.

Training on iSurg alone yields a substantial $+18.1$ Macro gain, underscoring large-scale surgical data as the foundation on which the methodological components build.
TSL contributes $+2.4$ Macro through its three terms.
The consistency loss gives the main gain ($+1.4$), as intra-track coherence benefits both instrument and tissue tracking. 
The repulsion loss adds a further improvement by keeping unreliable or absent-frame tokens away from the anchor, 
and the VLC objective completes TSL at $78.3$, where category-level instrument semantics strengthen the learned identity.
SLM adds a further $+2.7$ Macro through its two components. 
Long-range sampling raises the Macro to $79.1$ by training the memory attention to use distant references, strengthening the model's long-term temporal modeling.
Anchor-aligned selection then completes the full SurgSLOT model and reaches the best Macro of $81.0$, with noticeable gains concentrated on the long-duration subsets (EV17, RARP50, Hyst-YT, and NUH-Hyst), confirming that semantic-anchored long-term memory benefits tracking in long-duration procedures.
Together, the two components yield $+5.1$ Macro over iSurg fine-tuned SAM2.

\begin{table}[!tp]\centering
\caption{Inference-time memory strategies on the same SurgSLOT-SAM2 checkpoint, where only the selection rule differs.}
\label{tab:memory_strategy}
\footnotesize
\setlength{\tabcolsep}{6pt}
\begin{threeparttable}
\begin{tabular}{lccc}
\toprule[1pt]
Memory strategy & \makecell{Instrument\\Avg.} & \makecell{Tissue\\Avg.} & \makecell{Macro\\Avg.} \\
\midrule
Recent-frame (vanilla)$^\ast$    & 83.0 & 72.6 & 79.1 \\
\quad w/ extended memory         & 83.3 & 72.8 & 79.4 \\
\quad w/ confidence memory       & 84.0 & 73.1 & 79.9 \\
\quad w/ diversity-based memory  & 84.4 & 73.2 & 80.2 \\
\rowcolor{blue!11}
\quad w/ anchor-aligned memory   & \textbf{85.1} & \textbf{74.2} & \textbf{81.0} \\
\bottomrule[1pt]
\end{tabular}
\begin{tablenotes}[flushleft]
\footnotesize
\item[$^\ast$] Long-range sampling baseline with vanilla recent-frame memory.
\end{tablenotes}
\end{threeparttable}
\end{table}

\begin{table}[!tp]
\centering
\caption{Sensitivity analysis of SLM hyperparameters.}
\label{tab:sensitivity_analysis}
\begin{subtable}[t]{0.49\linewidth}
\centering
\caption{Distant block size $S_d$ during training.}
\label{tab:sparse_frames}
\footnotesize
\setlength{\tabcolsep}{3pt}
\begin{tabular}{ccc}\toprule[1pt]
$S_d$ & Macro Avg. & Train Time \\\midrule
2 & 80.5 & 36h \\
\rowcolor{blue!11}
3 & \textbf{81.0} & 40h \\
4 & 80.9 & 43h \\
5 & 80.7 & 46h \\
\bottomrule[1pt]
\end{tabular}
\end{subtable}
\hfill
\begin{subtable}[t]{0.49\linewidth}
\centering
\caption{Long-term memory capacity $N_l$ at inference.}
\label{tab:memory_capacity}
\footnotesize
\setlength{\tabcolsep}{4pt}
\begin{tabular}{ccc}\toprule[1pt]
$N_l$ & Macro Avg. & FPS \\\midrule
0 & 79.1 & \textbf{69}\\
2 & 80.5 & 68\\
\rowcolor{blue!11}
3 & \textbf{81.0} & 68 \\
5 & 80.8 & 65\\
7 & 80.9 & 63 \\
\bottomrule[1pt]
\end{tabular}
\end{subtable}
\end{table}

\subsubsection{Long-term Memory Strategy Analysis}
Having established the component-level gains in Table~\ref{tab:component_ablation}, we isolate the inference-time memory selection rule, the core of SLM.
Fixing the checkpoint trained in long-range sampling, we compare strategies for filling the long-term memory on the same model, so any difference is attributable to selection alone (Table~\ref{tab:memory_strategy}).
Simply extending the short-term memory by three frames yields $79.4$ Macro, as the added frames still hold recent content and provide little distant coverage. 
A confidence-based rule, selecting the highest-confidence frame within each span of consecutively high-confidence ($q_t > \gamma_q$) frames, reaches $79.9$ Macro, confirming that distant frames help but that confidence alone is a weak criterion. 
A diversity-based rule that selects frames for temporal coverage rather than redundancy, following ReSurgSAM2~\citep{resurgsam2}, reaches $80.2$ Macro, the best among these quality-driven rules, yet still falls short of identity-aware selection.
Anchor-aligned selection achieves the best $81.0$ Macro, with consistent gains on instruments and tissues, as the semantic anchor admits into long-term memory only frames consistent with the target identity, rejecting similar distractors. 
Overall, this progression shows that the gain comes specifically from selecting frames by semantic identity, not from storing more frames or from quality-based and diversity-driven selection rules, confirming that SLM depends on both trainable long-range sampling and identity-aware selection instead of acting as an inference-time heuristic.

\subsubsection{Analysis of Long-term Memory Configuration}
\label{sec:sensitivity}

We examine SurgSLOT's sensitivity to the two key SLM hyperparameters: the number of long-term frames seen during training and the long-term memory capacity used at inference, reported in Tables~\ref{tab:sparse_frames} and~\ref{tab:memory_capacity}.
Unless otherwise noted, SurgSLOT uses a distant block of $S_d=3$ and a recent block of $S_r=7$ in training, with long-term and short-term memory capacities $N_l=3$ and $N_s=6$ at inference.
Each hyperparameter is varied with the others held at these defaults, isolating its individual effect.

\noindent\textbf{Long-term Frames in Training.}
The distant block holds one conditional frame and $(S_d-1)$ long-term frames spread across long-range temporal gaps, training the memory attention for the long-range matching it performs at inference (Sec.~\ref{sec:slm_train}).
We start from $S_d=2$, since the shared temporal embedding receives gradients only when at least one frame enters the long-term memory.
Fixing the other hyperparameters at their defaults, accuracy peaks at $S_d=3$ ($81.0$) and then saturates, while training time keeps growing to $46$h at $S_d=5$.
A distant block of $S_d=3$, i.e., two long-term frames per clip, therefore suffices.

\noindent\textbf{Long-term Memory Capacity at Inference.}
Since long-term frames share one temporal embedding, $N_l$ can be varied without retraining. 
Fixing the training configuration at its defaults, disabling long-term memory ($N_l=0$) drops accuracy to $79.1$. 
Accuracy then peaks at $N_l=3$ ($81.0$) and saturates beyond it, while FPS falls from $68$ to $63$, showing that a compact set of identity-consistent frames outperforms storing more.

\begin{table}[!tp]\centering
\caption{Effect of training procedure diversity. Procedure types are added cumulatively to training.}
\label{tab:data_scale}
\footnotesize
\setlength{\tabcolsep}{8pt}
\begin{tabular}{lcc}
\toprule[1pt]
Training data & SAM2 & SurgSLOT-SAM2 \\
\midrule
Cholecystectomy   & 57.4 & 64.2 \\
+ Colonoscopy   & 62.7 & 67.8 \\
+ Gynecology   & 70.1 & 74.2 \\
+ Rectal Resection   & 73.5 & 77.3 \\
\rowcolor{blue!11}
+ Prostatectomy (Full)       & 75.9 & \textbf{81.0} \\
\bottomrule[1pt]
\end{tabular}
\end{table}

\subsubsection{Effect of Training Procedure Diversity}
\label{sec:data_scale}
To assess whether iSurg's procedural diversity is necessary, we train a separate model on each cumulative subset of procedure types, adding one type at a time, and evaluate every model on the full iSurg test set, reporting Macro Average $\mathcal{J}\&\mathcal{F}$ in Table~\ref{tab:data_scale}.
Both models improve steadily as more procedures are included, with SurgSLOT-SAM2 rising from $64.2$ with a single procedure to $81.0$ on the full iSurg, showing that broad procedural coverage drives cross-dataset generalization and that adding procedure types brings consistent gains.
SurgSLOT-SAM2 also outperforms SAM2 across all data scales, by $3.8$ to $6.8$ points, indicating that its gains come from the proposed design instead of training data alone.

\section{Discussion}
\label{sec:discussion}
A central design principle behind SurgSLOT is that a single learned semantic anchor plays two roles: during training, it shapes the target's representation; during inference, it selects which past frames to remember. 
This dual role distinguishes SurgSLOT from prior memory mechanisms, which rank frames by predicted quality and thus cannot separate the true target from an equally well-segmented distractor. 
Our ablation study confirms that identity-based selection, rather than storing more or higher-confidence frames, is what keeps the long-term memory reliable across procedures spanning tens of minutes.
The identity learned by TSL is also inherently suited to generalization: the label-free coherence objective forms an identity for any tracked target without binding it to a fixed taxonomy, while the vision-language contrastive learning organizes instrument tokens by transferable category semantics, and at inference, the anchor is accumulated online from the test video itself for memory selection in SLM.

Although anchor-aligned selection drives the long-term memory at inference, training instead uses long-range sampling, and we discuss this choice here. 
At inference, with no ground truth, frame reliability needs to be judged through anchor similarity. 
During training, the ground-truth masks reveal directly which frames are reliable and whether the target is present, so the anchor proxy is not needed. 
Anchor-aligned selection also runs online over the whole video, which would require keeping the whole video in memory during training and is too expensive, whereas long-range sampling draws only a few distant frames per clip, adding little overhead.
More importantly, random long-range sampling gives a wider training distribution than anchor-based selection, covering target-absent and visually varied frames, so the memory attention trains under more challenging conditions and generalizes better.

\begin{table}[!tp]\centering
\caption{Effect of input resolution on SAM2.}
\label{tab:resolution}
\footnotesize
\setlength{\tabcolsep}{8pt}
\begin{tabular}{lccc}\toprule[1pt]
Setting &Resolution &Macro Avg. &FPS \\\midrule
\multirow{2}{*}{Vanilla} 
  &1024 &64.2 &26 \\
  &512  &57.8 &69 \\ 
\cmidrule{1-4}
\multirow{2}{*}{iSurg FT} 
  &1024 &\textbf{76.2} &26 \\
  &512  &75.9 &\textbf{69} \\
\bottomrule[1pt]
\end{tabular}
\end{table}

Input resolution offers a practical way to trade off accuracy and speed in deployment.
As reported in Table~\ref{tab:resolution}, vanilla SAM2 depends on $1024$ resolution and falls to $57.8$ Macro at $512$, while after iSurg fine-tuning, the two resolutions become nearly equivalent ($76.2$ versus $75.9$) and $512$ runs $2.7\times$ faster ($69$ versus $26$ FPS). 
Once surgical fine-tuning closes the domain gap, resolution is no longer the bottleneck, which is why we adopt $512$ for all fine-tuned SAM2-based models and set SAM3 to $672$ to balance accuracy and efficiency.

Taken together, SurgSLOT's cross-dataset generalization and real-time efficiency carry practical implications for surgical deployment.
Collecting and annotating data for every new hospital, surgeon, or procedure type is a major cost barrier to deploying segmentation in the operating room. 
By generalizing across procedures and categories without per-context retraining, SurgSLOT can reduce this annotation burden.
Its $68$ FPS inference also runs in real time, fast enough for deployment in the operating room.

Despite these strengths, SurgSLOT still struggles with intrinsically difficult targets, such as tiny, frequently occluded structures (e.g., the \textit{suturing needle}) and highly deformable tissues, which remain challenging for all evaluated methods. 
Addressing them motivates two directions: tiny-structure segmentation through high-resolution or multi-scale feature refinement and boundary-aware supervision, and deformation-aware propagation through motion or non-rigid registration cues.

\section{Conclusion}
We address generalizable surgical video segmentation with the iSurg benchmark and SurgSLOT, a surgical segmentation generalist.
iSurg provides masklet annotations across six procedure types, with splits for cross-dataset and zero-shot evaluation.
SurgSLOT segments any prompted target via semantic long-term tracking, built on two coupled modules: TSL learns an object-level semantic identity for target re-identification, and SLM uses this identity to select reliable long-term memory frames and suppress drift across extended surgical videos.
Instantiated on both SAM2 and SAM3, SurgSLOT achieves state-of-the-art cross-dataset and zero-shot generalization, with its SAM2 version maintaining $68$ FPS for real-time inference.
Future work includes improving tiny-structure and deformable-tissue tracking, and extending the self-evolving data engine to cover more procedure types and datasets while reducing manual verification effort.

\printcredits

\section*{Declarations}
\noindent\textbf{Ethics Approval.}
This study was approved by the National Healthcare Group Domain Specific Review Board (NHG DSRB), National University Hospital, Singapore (Approval No.\ 2025-1189).

\noindent\textbf{Competing Interests.}
The authors declare that they have no known competing financial interests or personal relationships that could have appeared to influence the work reported in this paper.

\noindent\textbf{Data and Code Availability.}
The iSurg benchmark, source code, and pre-trained models will be made publicly available at \url{https://jinlab-imvr.github.io/SurgSLOT} upon acceptance of this manuscript.

\noindent\textbf{Use of AI-Assisted Technologies.}
During the preparation of this work, the authors used Claude Opus 4.8 to improve the language, readability, and logical flow of the manuscript. After using this tool, the authors reviewed and edited the content as needed and take full responsibility for the content of the published article.

\bibliographystyle{cas-model2-names}

\bibliography{references}

\end{document}